\pdfoutput=1

\documentclass[10pt,twocolumn,twoside]{IEEEtran}
\usepackage{color}
\usepackage{algorithmicx}
\usepackage[ruled]{algorithm}
\usepackage{algpseudocode}
\usepackage{url}
\usepackage{amsmath}
\usepackage{amssymb}
\usepackage{bbm}

\usepackage{subfigure}
\usepackage{graphicx} 

\DeclareMathOperator*{\argmin}{arg\,min}
\DeclareMathOperator*{\argmax}{arg\,max}
\graphicspath{{./}{./Figs/}}

\def\Phi{{\rm \psi }}

\begin{document}

\title{DeepTrack: Learning Discriminative Feature Representations Online for Robust Visual
Tracking}

\author{Hanxi Li, ~ Yi Li, ~ Fatih Porikli
\thanks
{
H. Li is with the School of Computer and Information Engineering, Jiangxi Normal
University, Jiangxi, China. He is also with NICTA, Canberra Research
Laboratory, Canberra, ACT2601, Australia,
(e-mail: hanxi.li@nicta.com.au).

Y. Li and F. Porikli are with NICTA, Canberra Research Laboratory, Canberra, ACT2601,
Australia, and also with the Research School of Information Science and Engineering,
Australian National University,
(e-mail: \{yi.li, fatih.porikli\}@nicta.com.au)

Correspondence should be addressed to Y. Li. 
}
}

\maketitle

\begin{abstract}

Deep neural networks, albeit their great success on feature learning in various computer
vision tasks, are usually considered as impractical for online visual tracking because
they require very long training time and a large number of training samples. In this work,
we present an efficient and very robust tracking algorithm using a single Convolutional
Neural Network (CNN) for learning effective feature representations of the target
object, in a purely online manner. Our contributions are multifold: 
First, we introduce a novel truncated structural loss function that maintains as many
training samples as possible and reduces the risk of tracking error accumulation. 
Second, we enhance the ordinary Stochastic Gradient Descent approach in CNN training with
a robust sample selection mechanism. The sampling mechanism randomly generates positive
and negative samples from different temporal distributions, which are generated by
taking the temporal relations and label noise into account.
Finally, a lazy yet effective updating scheme is designed for CNN training. Equipped with
this novel updating algorithm, the CNN model is robust to some long-existing difficulties
in visual tracking such as occlusion or incorrect detections, without loss of the
effective adaption for significant appearance changes. 
In the experiment, our CNN tracker outperforms all compared state-of-the-art methods on
two recently proposed benchmarks which in total involve over $60$ video sequences. The
remarkable performance improvement over the existing trackers illustrates the superiority
of the feature representations which are learned purely online via the proposed deep
learning framework.
\end{abstract}

\section{Introduction}

Image features play a crucial role in many challenging computer vision tasks such as
object recognition and detection. Unfortunately, in many \textit{online} visual trackers
features are manually defined and combined
\cite{perez2002color,10.1109/TPAMI.2005.205,adam2006robust,hare2011struck}.  Even though
these methods report satisfactory results on individual datasets, hand-crafted feature
representations would limit the performance of tracking. For instance, normalized cross
correlation, which would be discriminative when the lighting condition is favourable,
might become ineffective when the object moves under shadow.  This necessitates good
representation learning mechanisms for visual tracking that are capable of capturing the
appearance effectively changes over time.

Recently, deep neural networks have gained significant attention thanks to their success
on learning feature representations. Different from the traditional hand-crafted
features~\cite{lowe2004distinctive,dalal2005histograms,ahonen2006face}, a multi-layer
neural network architecture can efficiently capture sophisticated hierarchies describing
the raw data~\cite{Bengio2012}. In particular, the Convolutional Neural Networks (CNN) has
shown superior performance on standard object recognition
tasks~\cite{Kavukcuoglu2010,Krizhevsky2012,Ciresan2012b,girshick2014rich,zhang2014part}, which
effectively learn complicated mappings while utilizing minimal domain knowledge. 

However, the immediate adoption of CNN for online visual tracking is not straightforward.
First of all, CNN requires a large number of training samples, which is often not be
available in visual tracking as there exist only a few number of reliable positive
instances extracted from the initial frames. Moreover, CNN tends to easily overfit to the
most recent observation, {\it e.g.}, most recent instance dominating the model, which may
result in drift problem. Besides, CNN training is computationally intensive for online
visual tracking. Due to these difficulties, CNN has been treated as an offline feature
extraction step on predefined datasets~\cite{Fan:2010:HTU:1892525.1892533,NIPS2013_5192}
for tracking applications so far.

In this work, we propose a novel tracking algorithm using CNN to automatically learn the
most useful feature representations of particular target objects while overcoming the
above challenges. We employ a tracking-by-detection strategy -- a four-layer CNN model to
distinguish the target object from its surrounding background. Our CNN generates scores
for all possible hypotheses of the object locations (object states) in a given frame. The
hypothesis with the highest score is then selected as the prediction of the object state
in the current frame. We update this CNN model in an purely online manner. In other words,
the proposed tracker is learned based only on the video frames for the interested object,
no extra information or offline training is required. 

Typically, tracking-by-detection approaches rely on predefined heuristics to sample from
the estimated object location to construct a set of positive and negative samples. Often
these samples have binary labels, which leads to a few positive samples and a large
negative training set. However, it is well-known that CNN training without any pre-learned
model usually requires a large number of training samples, both for positive ones and
negative ones. Furthermore, even with sufficient samples, the learner usually needs
hundreds of seconds to achieve a CNN model with an acceptable accuracy. The slow updating
speed could prevent the CNN model from being a practical visual tracker.
%
%
To address these two issues, our CNN model employs a special type of loss function that
consists of a structural term and a truncated norm. The structural term
makes it possible to obtain a large number of training samples that have different
significance levels considering the uncertainty of the object location at the same time.
The truncated norm is applied on the CNN response to reduce the number of samples in the
back-propagation \cite{Kavukcuoglu2010,Krizhevsky2012} stage to significantly accelerate
the training process. 

We employ the Stochastic Gradient Decent (SGD) method to optimize the parameters in the
CNN model. Since the standard SGD algorithm is not tailored for online visual tracking, we
propose the following two modifications. First, to prevent the CNN model from overfitting
to occasionally detected false positive instances, we introduce a \emph{temporal sampling
mechanism} to the batch generation in the SGD algorithm. This temporal sampling mechanism
assumes that the object patches shall stay longer than those of the background in the
memory. Therefore, we store all the observed image patches into training sample pool, and
we choose the positive samples from a temporal range longer than the negative ones. In
practice, we found this is a key factor in the robust CNN-based tracker, because
discriminative sampling strategy successfully regularizes the training for effective
appearance model. Secondly, the object locations, except the one on the first frame, is
not always reliable as they are estimated by the visual tracker and the uncertainty is
unavoidable \cite{babenko2009visual}. One can treat this difficulty as the label noise
problem \cite{lawrence2001estimating, long2010random, natarajan2013learning}. We propose
to sample the training data in the joint distribution over the temporal variable (frame
index) and the sample class. Here we compute the conditional probability of the sample
class, given the frame index, based on a novel measurement of the tracking quality in that
frame. In the experiment, further performance improvement is observed when the sample
class probability is taken into account.

For achieving a high generalization ability in various image conditions, 
we use multiple image \emph{cues} (low-level image features, such as
normalized gray-scale image and image gradient) as independent channels as network input.
We update the CNN parameters by iteratively training each channel independently followed
by a joint training on a fusion layer which replace the last fully-connected layers from
multiple channels. The training processes of the independent channels and the fusion layer
are totally decoupled. This makes the training efficient and empirically we observed that
this two-stage iterative procedure is more accurate than jointly training for all cues. 

Finally, we propose to update the CNN model in a ``lazy'' style. First, the CNN-model is
only updated when a significant appearance change occurs on the object. The intuition
behind this lazy updating strategy is that we assume that the object appearance is more
consistent over the video, compared with the background appearances. Second, the fusion
layer is updated in a coordinate-descent style and with a lower learning rate. The
underlying assumption is that the feature representations can be updated fast while the
contribution ratios of different image cues are more stable over all the frames.
In practice, this lazy updating strategy not only increases the tracking speed significantly
but also yields observable accuracy increase. 

To summarize, our main contributions include:
\begin{itemize}
\item A visual tracker based on online adapting CNN is proposed. As far as we are aware,
  this is the first time a single CNN is introduced for learning the best features for
  object tracking in an online manner. 

\item A structural and truncated loss function is exploited for the online CNN
  tracker. This enables us to achieve very reliable (best reported results in the
  literature) and robust tracking while achieving tracking speeds up to $4$fps.

\item An iterative SGD method with an robust temporal sampling mechanism is introduced for
  competently capturing object appearance changes and meanwhile considering the label noise.

\end{itemize}

Our experiments on two recently proposed benchmarks involving over $60$ videos demonstrate
that our method outperforms all the compared state-of-the-art algorithms and rarely loses
the track of the objects. In addition, it achieves a practical tracking speed (from
$1.5$fps to $4$fps depending on the sequence and settings), which is comparable to many
other visual trackers.

\section{CNN Architecture}
\label{sec:architecture}

\begin{figure*}[t]
\begin{centering}
\includegraphics[width=0.95\textwidth]{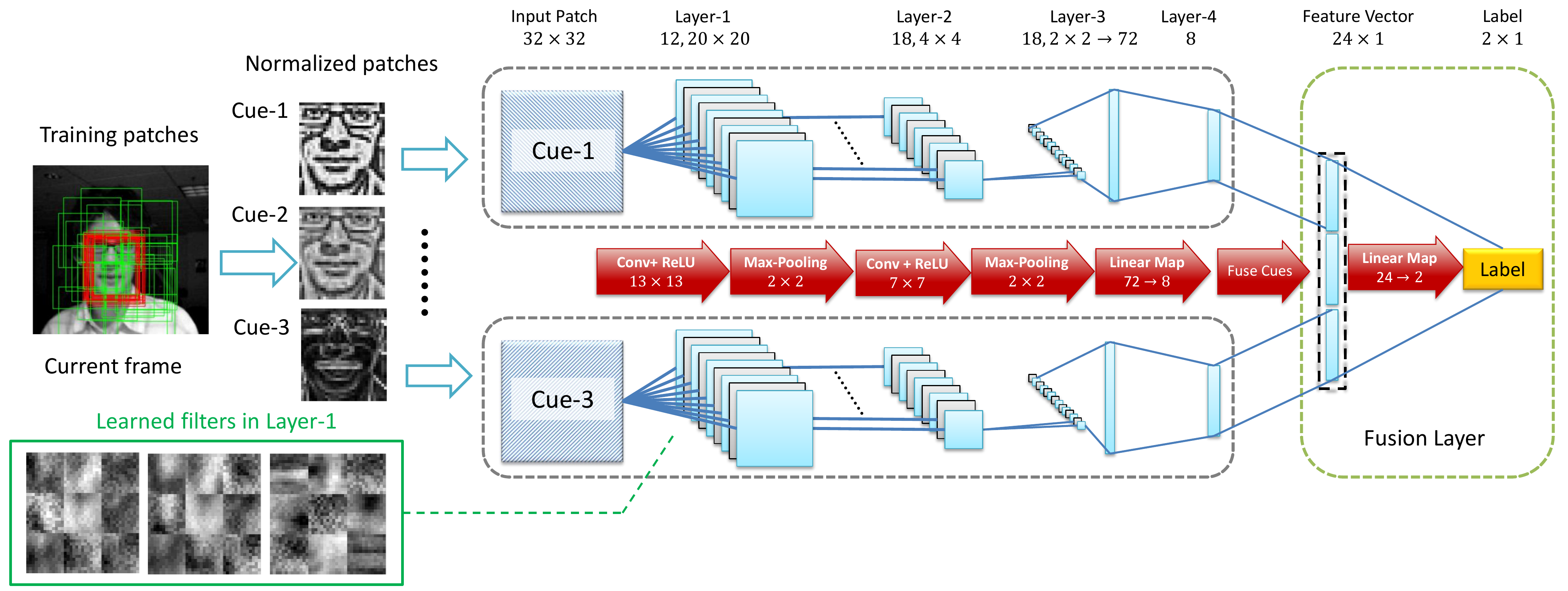}
\par\end{centering}
\caption{
    The architecture of our CNN tracker with multiple image cues. The gray dashed blocks
    are the independent CNN channels for different image cues; the green dashed block is
    the fusion layer where a linear mapping $\mathbb{R}^{24}\to\mathbb{R}^{2}$ is learned.
}  
\label{fig:netOverview}
\end{figure*}

\subsection{CNN with multiple image cues}

Our CNN consists of two convolutional layers and two fully-connected layers. The ReLU
(Rectified Linear Unit) \cite{Krizhevsky2012} is adopted as the activation function and
max-pooling operators are used for dimension-reduction. 
The dark gray block in Fig.~\ref{fig:netOverview} shows
the structure of our network, which can be expressed as
$(32\times32)\to(10\times10\times12)\to(2\times2\times18)\to(8)\to(2)$ in conventional
neural network notation. 

The input is locally normalized $32\times32$ image patches, which draws a balance between
the representation power and computational load. The first convolution layer contains $12$
kernels each of size $13\times13$ (an empirical trade-off between overfitting due to a
very large number of kernels and discrimination power), followed by a pooling operation
that reduces the obtained feature map (filter response) to a lower dimension. The second
layer contains $216$ kernels with size $7\times7$. This leads to a $72$-dimensional
feature vector in the second convolutional layer, after the pooling operation in this
layer. 

The two fully connected layers firstly map the $72$-D vector into a $8$-D vector and then
generate a $2$-D confidence vector $\bold{s} = [s_1, s_2]^{\text{T}} \in \mathcal{R}^2$,
with $s_1$ and $s_2$ corresponding to the positive score and negative score, respectively.
In order to increase the margin between the scores of the positive and negative samples,
we calculate the CNN score of the patch $n$ as 
\begin{equation}
  S(\bold{x}_n;\Omega) = S_n = s_1\cdot\text{exp}(s_1 - s_2), 
  \label{equ:cnn_score}
\end{equation}
where $\bold{x}_n$ denotes the input and the CNN is parameterized by the weights $\Omega$.

Effective object tracking requires multiple cues, which may include color, image gradients
and different pixel-wise filter responses. These cues are weakly correlated yet contain
complementary information. 
Local contrast normalized cues are previously shown \cite{Krizhevsky2012} to produce
accurate object detection and recognition results within the CNN frameworks. The
normalization not only alleviates the saturation problem but also makes the CNN robust to
illumination change, which is desired during the tracking.  In this work, we use $3$ image
cues generated from the given gray-scale image, {\it i.e.}, two locally normalized images
with different parameter configurations \footnote{Two parameters $r_{\mu}$ and
  $r_{\sigma}$ determine a local contrast normalization process.  In this work, we use two
configurations, \emph{i.e.}, $\{r_{\mu} = 8, r_{\sigma} = 8\}$ and $\{r_{\mu} = 12,
r_{\sigma} = 12\}$, respectively.} and a gradient image. For color images, the first two cues are simply
replaced with the H and V channels of the HSV color representation. Offering multiple
image cues, we then let CNN to select the most informative ones in a data driven fashion. By
concatenating the final responses of these $3$ cues, we build a fusion layer (the green
dashed block in Fig.~\ref{fig:netOverview}) to generate a $2$-D output vector, based on
which the final CNN score is calculated using Eq.~\ref{equ:cnn_score}. 

In our previous work \cite{ourBMVC2014, ourACCV2014}, we proposed to use a set of CNNs
\cite{ourBMVC2014}, or a single CNN \cite{ourACCV2014} with multiple ($4$) image cues for
visual tracking. In this work, we employ a more complex CNN model (as described above) while
less image cues to strike the balance between robustness and tracking speed. Other
small yet important modifications from the previous model includes: 
\begin{itemize}
  \item To better curb the overfitting, all the training samples are flipped
    as augmented data.
  \item The pixel values of each the image cue are normalized to the range $[0, 10]$. We
    found this normalization is crucial for balancing the importances between different
    image cues.
\end{itemize}

\subsection{Structural and truncated loss function}\label{s:loss}

\subsubsection{Structural loss}

Let $\bold{x}_n$ and $\bold{l}_n \in \{[0, 1]^{\text{T}}, [1, 0]^{\text{T}}\}$ denote the
cue of the input patch and its ground truth label (background or foreground) respectively,
and $f(\bold{x}_n;\Omega)$ be the predicted score of $\bold{x}_n$ with network weights
$\Omega$, the objective function of $N$ samples in the batch is
\begin{equation}
  \mathcal{L} = \frac{1}{N}\displaystyle\sum_{n=1}^{N}\left \| f(\bold{x}_n;\Omega)-\bold{l}_n \right \|_{2}
\label{equ:normal_loss}
\end{equation}
when the CNN is trained in the batch-mode. Eq.~\ref{equ:normal_loss} is a commonly
used loss function and performs well in binary classification problems. However, for
object localization tasks, usually higher performance can be obtained  by `structurizing'
the binary classifier. The advantage of employing the structural loss is the larger number
of available training samples, which is crucial to the CNN training. In the ordinary
binary-classification setting, one can only use the training samples with high confidences
to avoid class ambiguity. In contrast, the structural CNN is learned based upon all the
sampled patches.

We modify the original CNN's output to $f(\phi\langle\Gamma,\bold{y}_n\rangle;\Omega) \in
\mathbb{R}^2$, where $\Gamma$ is the current frame, $\bold{y}_n \in \mathbb{R}^{o}$ is the
motion parameter vector of the target object, which determines the object's location in
$\Gamma$ and $o$ is the freedom degree\footnote{In this paper $o = 3$, {\it i.e.}, the
bounding box changes in its location and the scale.} of the transformation. The operation
$\phi\langle\Gamma,\bold{y}_n\rangle$ suffices to crop the features from $\Gamma$ using
the motion $\bold{y}_n$.  The associated structural loss is defined as
\begin{equation}
  \mathcal{L} = \frac{1}{N}\displaystyle\sum_{n=1}^{N}
  \left[\Delta(\bold{y}_n, \bold{y}^{\ast})\cdot\left\|
  f(\phi\langle\Gamma,\bold{y}_n\rangle;\Omega) - \bold{l}_n \right \|_{2}\right],
  \label{equ:structural_loss}
\end{equation}
where $\bold{y}^{\ast}$ is the (estimated) motion state of the target object in the
current frame. To define $\Delta(\bold{y}_n, \bold{y}^{\ast})$ we first calculate the
overlapping score $\Theta(\bold{y}_n, \bold{y}^{\ast})$ \cite{everingham2010pascal} as
\begin{equation}
  \Theta(\bold{y}_n, \bold{y}^{\ast}) = \frac{\text{area}(r(\bold{y}_n) \bigcap
  r(\bold{y}^{\ast}))}{\text{area}(r(\bold{y}_n) \bigcup r(\bold{y}^{\ast}))}
  \label{equ:overlap_loss}
\end{equation} 
where $r(\bold{y})$ is the region defined by $\bold{y}$, $\bigcap$ and $\bigcup$ denotes
the intersection and union operations respectively. Finally we have
\begin{equation}
  \Delta(\bold{y}_n, \bold{y}^{\ast}) = \left|\frac{2}{1 +
  \text{exp}(-(\Theta(\bold{y}_n, \bold{y}^{\ast}) - 0.5))} - 1\right| \in
  [0, 1].
\label{equ:sigmoid_overlap}
\end{equation}
And the sample label $\bold{l}_n$ is set as.
\[ \bold{l}_n = \left\{ 
  \begin{array}{l l}
  [1, 0]^{\text{T}} & \quad \text{if} \quad \Theta(\bold{y}_n, \bold{y}^{\ast}) > 0.5 \\
{[0, 1]^{\text{T}}} & \quad \text{elsewise}
  \end{array} \right.\]
From Eq. \ref{equ:sigmoid_overlap} we can see that $\Delta(\bold{y}_n, \bold{y}^{\ast})$
actually measures the importance of the training patch $n$. For instance, patches
that are very close to object center and reasonably far from it may play more significant
roles in training the CNN, while the patches in between are less important. 

In visual tracking, when a new frame $\Gamma_{(t)}$ comes, we predict the object motion
state $\bold{y}_{(t)}^{\ast}$ as
\begin{equation}
  \bold{y}^{\ast}_{(t)} = \argmax_{\bold{y}_n \in
  \mathcal{Y}}\left(f(\phi\langle\Gamma_{(t)},\bold{y}_n\rangle;\Omega)\right),
\label{equ:tracking_infer}
\end{equation}
where $\mathcal{Y}$ contains all the test patches in the current frame. 

\subsubsection{Truncated structural loss}

Ordinary CNN models regress the input features into the target labels, via the 
$l_2$-norm loss. One can directly adopt this strategy in the CNN-based tracking
algorithm. However, to speed up the online training process, we employ a truncated
$l_2$-norm in our model. We empirically observe that patches with very small error does
not contribute much in the back propagation. Therefore, we can approximate the loss by
counting the patches with errors that are larger than a threshold. Motived by this, in
\cite{ourACCV2014}, we define a truncated $l_2$ norm as 
\begin{equation}
  \|e\|_{\mathbb{T}} = \|e\|_2 \cdot \left(1 - \mathbbm{1}[\|e\|_2 \le \beta]\right),
  \label{equ:truncated_l2}
\end{equation}
where $\mathbbm{1}[\cdot]$ denotes the indicator function while $e$ is the prediction
error, \emph{e.g.}, $f(\phi\langle\Gamma,\bold{y}_n\rangle;\Omega) - \bold{l}_n$ for
patch-$n$. In our previous work \cite{ourACCV2014}, this truncated loss did increase the
training speed, while at the cost of reducing the prediction accuracy. 

\begin{figure}[h]
\centering
\includegraphics[width=0.35\textwidth]{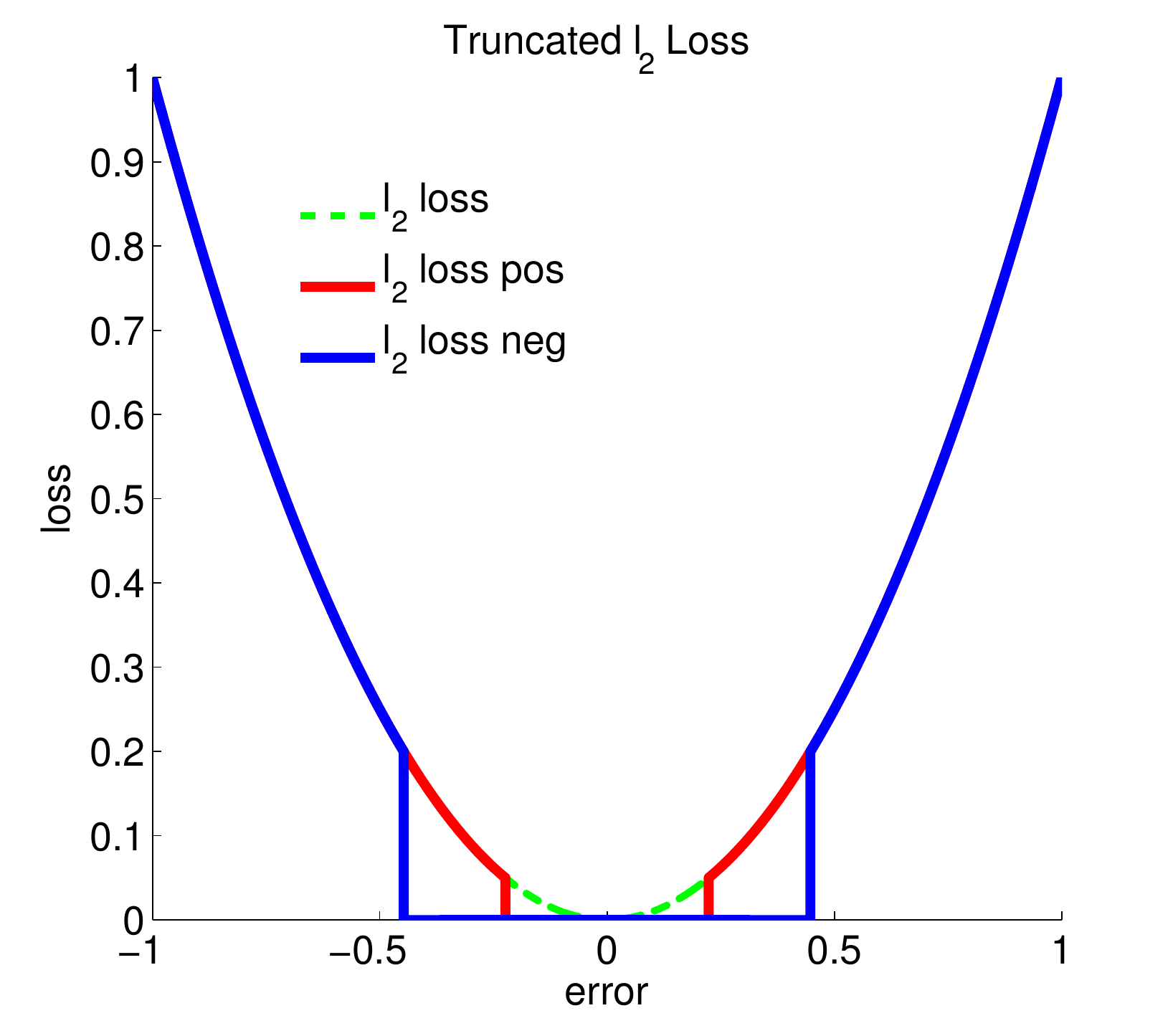}
\caption{
  The truncated $l_2$ losses. The dashed green curve indicates the original $l_2$ loss,
  the red and blue curves are the truncated losses for positive and negative samples.
}  
\label{fig:truncated_loss}
\end{figure}

In this work, we observed that the tracking performance is more sensitive to the
prediction error on positive samples than the negative samples. Recall that in training
stage, we label each positive sample as $[1, 0]^{T}$ and each negative sample as $[0,
1]^{T}$. In the test stage, the visual tracker selects the best particle among the ones
with high scores. If the highest score in the current frame is large enough, the negative
samples with small errors, which are ignored in training according to the truncated loss,
will not affect the prediction. In contrast, if one ignores the positive samples with
small errors in training, the selection among the top-$n$ particles in the test stage will
be consequently inaccurate, and thus drift problems could happen. 
In other words, we need a more precise loss function for
positive samples in visual tracking. We thus improve the original truncated loss function
as: 
\begin{equation}
  \|e\|_{\mathbb{T}} = \|e\|_2 \cdot \left(1 - \mathbbm{1}\left[\|e\|_2 \le \frac{\beta}{(1 +
  u \cdot l_n)}\right]\right),
  \label{equ:truncated_l2_new}
\end{equation}
where $u > 0$ and $l_n = \bold{l}_n(1)$, \emph{i.e.}, the scalar label of the $n$-th
sample. 
This truncated norm is visualized in Fig.~\ref{fig:truncated_loss} and now
Eq.~\ref{equ:structural_loss} becomes:
\begin{equation}
  \mathcal{L} = \frac{1}{N}\displaystyle\sum_{n=1}^{N}
  \left[\Delta(\bold{y}_n, \bold{y}^{\ast})\cdot\left\|
  f(\phi\langle\Gamma,\bold{y}_n\rangle;\Omega) - \bold{l}_n \right \|_{\mathbb{T}}\right],
  \label{equ:truncated_structural_loss}
\end{equation}

It is easy to see that with the truncated norm $\|\cdot\|_{\mathbb{T}}$, the backpropagation
\cite{Kavukcuoglu2010} process only depends on the training samples with large errors,
\emph{i.e.}, $\|f(\phi\langle\Gamma,\bold{y}_n\rangle;\Omega) - \bold{l}_n\|_{\mathbb{T}}
> 0$. Accordingly, we can ignore the samples with small errors and the backpropagation
procedure is significantly accelerated. In this work, we use $\beta = 0.0025$ and $u = 3$.

\section{Optimization of CNN for Tracking}
\label{sec:opt}

\subsection{Online Learning: Iterative SGD with Temporal Sampling}
\label{subsec:onlinetrain}

\subsubsection{Temporal Sampling}
\label{subsubsec:temporal}

Following other CNN-based approaches \cite{Kavukcuoglu2010,Krizhevsky2012}, we used
Stochastic Gradient Decent (SGD) for the learning of the parameters $\Omega$. However, the
SGD we employ is specifically tailored for visual tracking. 

Different from detection and recognition tasks, the training sample pool grows gradually
as new frames come in visual tracking. Moreover, it is desired to learn a consistent
object model over {\it all} the previous frames and then use it to distinguish the object
from the background in the {\it current} frame. This implies that we can effectively learn
a discriminative model on a long-term positive set and a short-term negative set. 

Based on this intuition, we tailor the SGD method by embedding in a temporal sampling
process. In particular, given that the positive sample pool is $\mathbb{Y}^{+}_{1 : t} =
\{\bold{y}^{+}_{1, (1)}, \bold{y}^{+}_{2, (1)}, \dots, \bold{y}^{+}_{N-1, (t)},
\bold{y}^{+}_{N, (t)}\}$\footnote{Here we slightly abuse the notation of $\bold{y}$,
which denotes the motion state in the previous section. Here $\bold{y}$ indicates the
cropped image patch according to the motion state.} and the negative sample pool is
$\mathbb{Y}^{-}_{1 : t} = \{\bold{y}^{-}_{1, (1)}, \bold{y}^{-}_{2, (1)}, \dots,
\bold{y}^{-}_{N-1, (t)}, \bold{y}^{-}_{N, (t)}\}$, when generating a mini-batch for SGD,
we sample the positive pool with the probability 
\begin{equation}
  \text{Prob}(\bold{y}^{+}_{n, (t')}) = \frac{1}{tN}\quad, 
  \label{equ:pos_sample}
\end{equation}
while sample the negative samples with the probability
\begin{equation}
  \text{Prob}(\bold{y}^{-}_{n, (t')}) = \frac{1}{Z}\text{exp}\left[-\sigma(t -
  t')^2\right],
  \label{equ:neg_sample}
\end{equation}
where $\frac{1}{Z}$ is the normalization term and we use $\sigma = 10$ in this work. 

In a way, the above temporal selection mechanism can be considered to be similar to the
``multiple-lifespan'' data sampling~\cite{xing2013robust}. However, \cite{xing2013robust}
builds three different codebooks, each corresponding to a different lifespan, while we
learn one discriminative model based on two different sampling distributions.

\subsubsection{Robust Temporal Sampling with Label Noise}

In most tracking-by-detection strategy, the detected object $\bold{y}_{(t)}^{\ast}$ is
treated as a true-positive in the following training stage. However, among all the motion
states $\bold{y}_{(t)}^{\ast}, ~\forall t = 1, 2, \dots, T$, only the first one
$\bold{y}_{(1)}^{\ast}$ is always reliable as it is manually defined. Other motion states
are estimated based on the previous observations. Thus, the uncertainty of the prediction
$\bold{y}_{(t)}, ~\forall t > 1$ is usually unavoidable \cite{babenko2009visual}.  Recall
that, the structural loss defined in Eq.~\ref{equ:overlap_loss} could change
significantly if a minor perturbation is imposed on $\bold{y}_{(t)}$, one requires a
accurate $\bold{y}_{(t)}$ in every frame, which is, unfortunately, not feasible. 

In our previous work \cite{ourACCV2014}, we take the uncertainty into account by imposing
a robust term on the loss function \ref{equ:truncated_structural_loss}. The robust term is
designed in the principle of Multiple-Instance-Learning \cite{393,viola2005multiple} and
it alleviates over-fittings in some scenarios. However, the positive-sample-bag
\cite{ourACCV2014} could also reduce the learning effectiveness as it will confuse the
learner when two distinct samples are involved in one bag. Actually, other MIL-based
trackers also suffer from this problem \cite{babenko2009visual,393}. 

In this work, we propose a much simpler scheme for addressing the issue of prediction
uncertainty. Specifically, the prediction uncertainty is casted as a label noise problem
\cite{lawrence2001estimating, long2010random, natarajan2013learning}. We assume there
exist some frames, on which the detected ``objects'' are false-positive samples. In other words, the
some sample labels in $\mathbb{Y}^{+}_{1 : t}$ and $\mathbb{Y}^{-}_{1 : t}$ are
contaminated (flipped in the binary case). In the context of temporal sampling, the
assumption introduces an extra random variable $\boldsymbol{\eta}$ which represent the
event that the label is true ($\boldsymbol{\eta} = 1$) or not
($\boldsymbol{\eta} = 0$). The sampling process is now conduct in the joint
probability space $\{n = 1, 2, \cdots, N\} \times \{t' = 1, 2, \cdots, t\} \times
\{\boldsymbol{\eta} = 1, 0\}$ and the joint probability is 
\begin{equation}
  \text{Prob}(\bold{y}^{\pm}_{n, (t')}, \boldsymbol{\eta} = 1),
  \label{equ:joint_prob}
\end{equation}
where $\bold{y}^{\pm}_{n, (t')}$ stands for the selection of the $n$-th positive/negative
sample in the $t'$-th frame. According to the chain-rule, we have
\begin{equation}
  \begin{split}
  \text{Prob}(\bold{y}^{\pm}_{n, (t')}, & \boldsymbol{\eta} = 1)
  = \text{Prob}(t', n, \boldsymbol{\eta} = 1) \\
  & = \text{Prob}(\boldsymbol{\eta} = 1 \mid t', n)\cdot\text{Prob}(t', n) \\
  & = \text{Prob}(\boldsymbol{\eta} = 1 \mid t',
  n)\cdot\text{Prob}\left(\bold{y}^{\pm}_{n, (t')}\right) \\
  \end{split}
  \label{equ:chain_rule}
\end{equation}
where $\text{Prob}\left(\bold{y}^{\pm}_{n, (t')}\right)$ is given in
Eq.~\ref{equ:pos_sample} and \ref{equ:neg_sample} while the conditional probability
$\text{Prob}(\boldsymbol{\eta} = 1 \mid t', n)$ reflects the likelihood that the label of
sample $\bold{y}^{\pm}_{n, (t')}$ is not contaminated. 

\begin{figure*}[htb]
  \centering
  \subfigure{\includegraphics[width=0.27\textwidth]{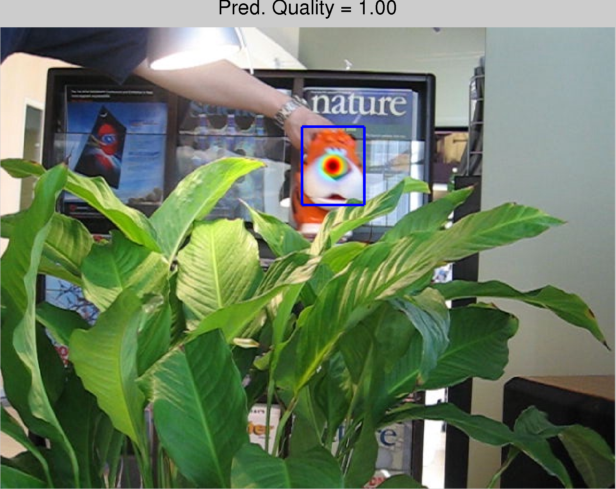}}\quad
  \subfigure{\includegraphics[width=0.27\textwidth]{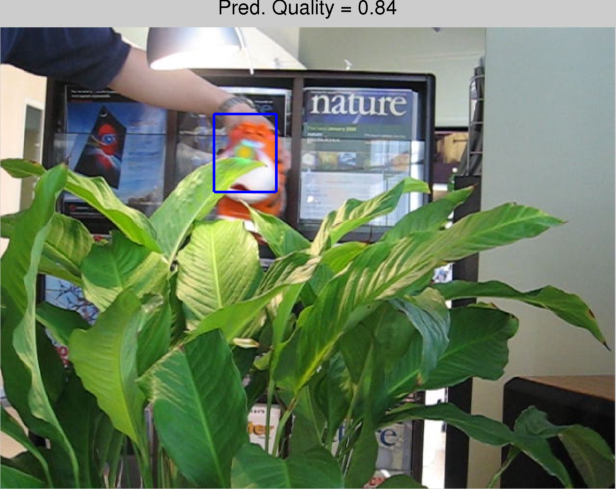}}\quad
  \subfigure{\includegraphics[width=0.27\textwidth]{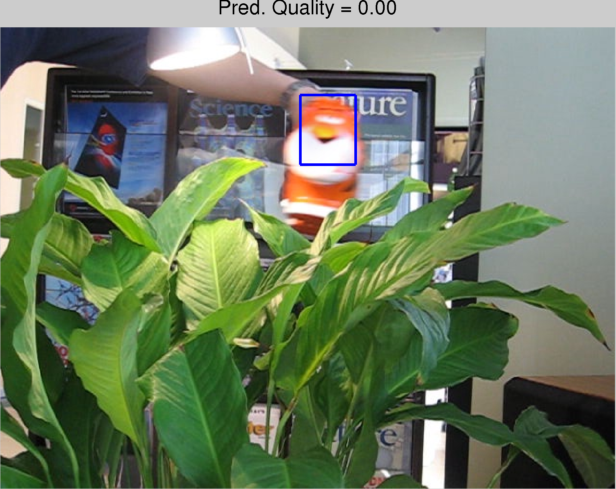}}
  \caption
  {
    A demonstration of the prediction quality on three different frames in the sequence
    \emph{tiger1}. For each frame, the overlaying heat-map indicates the distribution of the
    high-score particles while the blue box is the detected $\bold{y}^{\ast}_{(t')}$. The
    tracking qualities are shown on the top of the frame images. Note that the quality is
    estimated without ground-truth information.
  }
\label{fig:pred_quality}
\end{figure*}

To estimate $\text{Prob}(\boldsymbol{\eta} = 1 \mid t', n)$ efficiently, we assume that in
the same frame, the conditional probabilities are equal for $\forall n = 1, 2, \cdots, N$.
Then we propose to calculate the probability as a prediction quality $\text{Q}_{t'}$ in
frame-$t'$, \emph{i.e.},
\begin{equation}
  \begin{split}
    &\quad \text{Q}_{t'} = \text{Prob}(\boldsymbol{\eta} = 1 \mid t', n) = \\
    & 1 - \frac{1}{|\mathbb{P}|}\displaystyle\sum_{n \in \mathbb{P}}^{N}
      \left[\Delta(\bold{y}_{n,(t')}, \bold{y}^{\ast}_{(t')})\cdot\left\|
    f(\phi\langle\Gamma,\bold{y}_{n,(t')}\rangle;\Omega) - \bold{l}_n \right \|_{\mathbb{T}}\right],
  \end{split}
\label{equ:frame_conf}
\end{equation}
where the set $\mathbb{P}$ contains the sample in the frame $t'$ with high scores.
Mathematically, it is defined as 
\begin{equation}
  \mathbb{P} = \{\forall n \mid S_{n,(t')} > v\cdot S^{\ast}_{(t')}\},
  \label{equ:peak_set}
\end{equation}
where $S_{n,(t')}$ and $S^{\ast}_{(t')}$ are the CNN scores (see
Eq.~\ref{equ:cnn_score}) of the $n$-th sample and the sample selected as object in
frame $t'$, respectively. The underlying assumption of Eq.~\ref{equ:frame_conf} is
that, a detection heat-map with multiple widely-distributed peaks usually implies low
detection quality, as there is only ONE target in the video sequence.  This
tracking quality is illustrated in Fig.~\ref{fig:pred_quality}. From the figure we can
see that when occlusion (middle) or significant appearance change (right) occurs, the
tracking quality drops dramatically and thus the samples in those ``contaminated'' frames
are rarely selected according to Eq.~\ref{equ:chain_rule}.

\subsubsection{Iterative Stochastic Gradient Descent (IT-SGD)}

Recall that we use multiple image cues as the input of the CNN tracker. This leads to a
CNN with higher complexity, which implies a low training speed and a high possibility of
overfitting. By noticing that each image cue may be weakly independent, we train the
network in a iterative manner. In particular, we define the model parameters as $\Omega =
\{\mathbf{w}^1_{cov}, \cdots, \mathbf{w}^{K}_{cov}, \mathbf{w}^1_{fc}, \cdots,
\mathbf{w}^K_{fc}, \mathbf{w}_{fuse}\}$, where $\mathbf{w}^k_{cov}$ denotes the filter
parameters in cue-$k$, $\mathbf{w}^k_{fc}$ corresponds to the fully-connected layers and
$\mathbf{w}_{fuse}$ parameterize the fusion layer. 

In this work, we conduct the SGD process iteratively over different image cues and the
fusion layer. In specific, after we complete the training on $\mathbf{w}^k_{cov}$ and
$\mathbf{w}^k_{fc}$, we evaluate the filter responses from that cue in the last
fully-connected layer and then update $\mathbf{w}_{fuse}$ on the dimensions corresponding
to cue-$k$ . This can be regarded as a coordinate-descent variation of SGD.  In practice,
we found out both the robust temporal sampling mechanism and the IT-SGD significantly curb
the overfitting problem. The iterative SGD is illustrated in
Algorithm~\ref{alg:coordinate_descent}.

\alglanguage{pseudocode} 
\begin{algorithm}[h] 
\small 
\caption{Iterative SGD with robust temporal sampling}\label{Algorithm:ff}
\begin{algorithmic}[1] 
  \State {\bf Inputs:} Frame image $\Gamma_{(t)}$; Two sample pools $\mathbb{Y}^{+}_{1 : t}$,
  $\mathbb{Y}^{-}_{1 : t}$; \\
  Old CNN model ($K$ cues) $f_0(\phi\langle\Gamma_{(t)},\cdot\rangle;\Omega)$. \\
  Estimated/given $\bold{y}^{\ast}_{(t)}$; \\
  Learning rates $r$; $\hat{r}$; minimal loss $\varepsilon$;
  training step budget $M$.
\Procedure {IT-SGD}{$\mathbb{Y}^{+}_{1 : t}$, $\mathbb{Y}^{-}_{1 : t}$, $f$,
$\bold{y}^{\ast}$, $\hat{r}$, $r$, $M$}
\State Selected samples $\{\bold{y}_{1,(t)}, \bold{y}_{2,(t)}, \dots, \bold{y}_{N,(t)}\}$.
\State Generate associated labels $\bold{l}_{1,(t)}, \cdots, \bold{l}_{N,(t)}$ according to $\bold{y}^{\ast}_{(t)}$.
\State Estimate the prediction quality $\text{Q}_{t}$. 
\State Save the current samples and labels into $\mathbb{Y}^{+}_{1 : t}$ and $\mathbb{Y}^{-}_{1 : t}$.
\State Sample training instances according to $\text{Prob}(\bold{y}^{\pm}_{n, (t)}, \boldsymbol{\eta} = 1)$.
\For{$m\gets 0,~M-1$}
  \State $\mathcal{L}_m = \frac{1}{N}\displaystyle\sum_{n=1}^{N}
    \left[\Delta(\bold{y}_n, \bold{y}^{\ast})\cdot\left\|
    f_m(\phi\langle\Gamma_{(t)},\bold{y}_n\rangle;\Omega) - \bold{l}_n \right
      \|_{\mathbb{T}}\right]$;
  \State {\bf If} {$\mathcal{L}_m \le \varepsilon$}, {\bf break};
  \State $k = \text{mod}(m, K) + 1$;
  \State Update $\mathbf{w}^k_{cov}$ and $\mathbf{w}^k_{fc}$ using SGD with learning rate $r$.
  \State Update $\mathbf{w}_{fuse}$ partially for cue-$k$, with learning rate $\hat{r}$.
  \State Save $f_{m+1} = f_m$;
\EndFor
\EndProcedure
\State {\bf Outputs:} New CNN model $f^{\ast} = f_{m^{\ast}}, m^{\ast} = \argmin_{m}\mathcal{L}_m$.
\Statex 
\end{algorithmic}   
\label{alg:coordinate_descent}
\end{algorithm}

\subsection{Lazy Update and the Overall Work Flow}

It is straightforward to update the CNN model using the IT-SGD algorithm at each frame.
However, this could be computationally expensive as the complexity of training processes
would dominate the complexity of the whole algorithm. On the other hand, in case the
appearance of the object is not always changing, a well-learned appearance model can
remain discriminant for a long time. Furthermore, when the feature representations is
updated for adapting the appearance changes, the contribution ratios 
of different image cues could remain more stable over all
the frames.

\begin{figure*}[htb]
  \centering
  \includegraphics[width=0.9\textwidth]{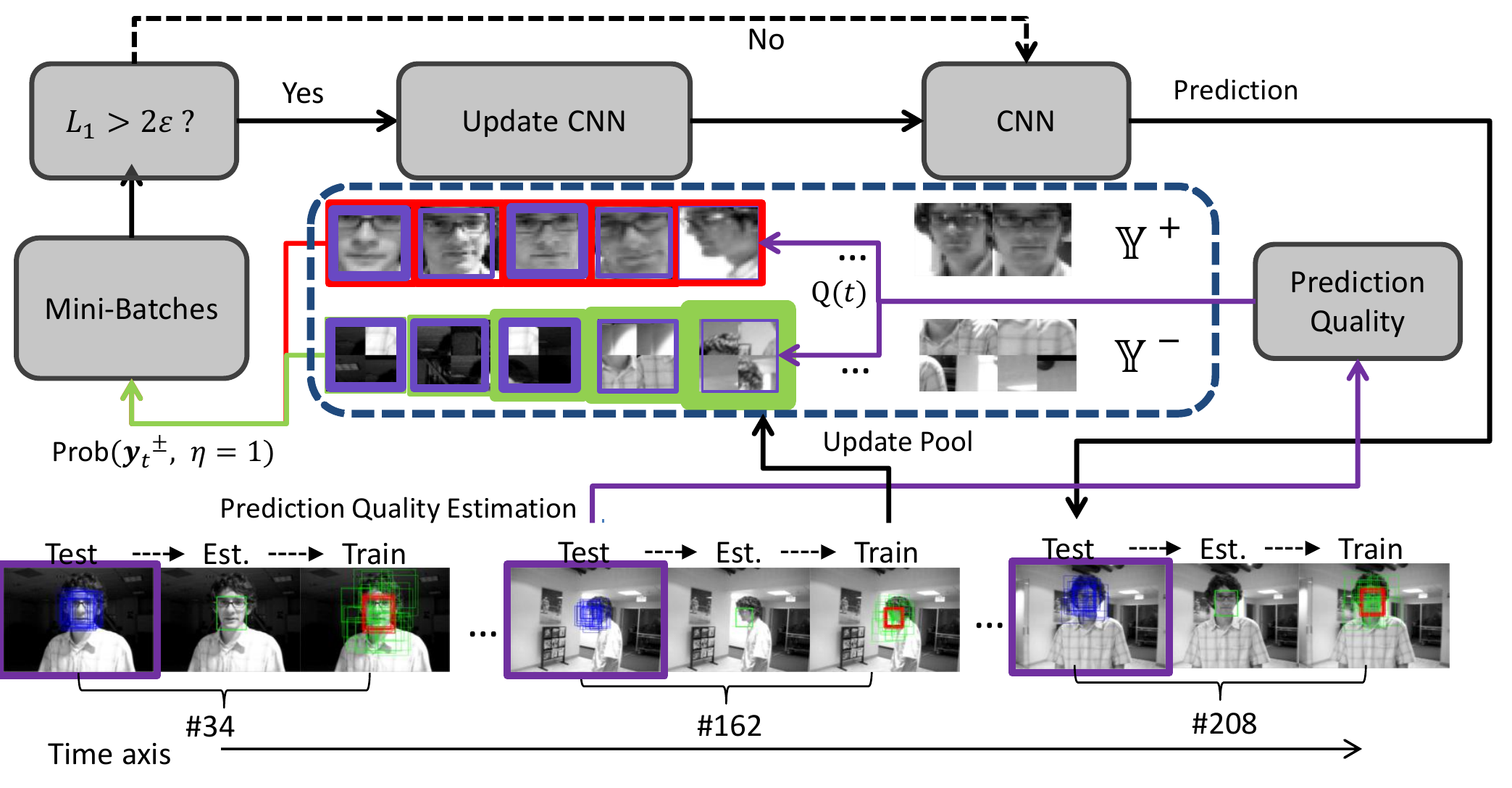}
  \caption
  {
    Work flow of proposed algorithm. The bottom row shows the three-stages operations on a
    frame: test, estimation and training. In the training frames, the green bounding-boxes
    are the negative samples while the red ones denote the positive samples. The dashed
    block covers the positive sample pool $\mathbb{Y}^{+}$ (red) and negative sample pool
    $\mathbb{Y}^{-}$ (green). In each pool, the edges of the sample patches indicate their sampling
    importances. The green ones (negative) and red ones (positive) represent the prior
    probabilities of sample selection while the purple ones stands for the conditional
    probabilities ($\text{Q}(t)$). The thicker the edge, the higher the probability.}
\label{fig:temporal_modeling}%
\end{figure*}

Motivated by the above two intuitions, we propose to update the CNN model in a lazy
manner. First, when tracking the object, we only update the CNN model when the training
loss $\mathcal{L}_1$ is above $2\varepsilon$. Once the training start, the training goal
is to reduce $\mathcal{L}$ below $\varepsilon$. As a result, usually $\mathcal{L}_1 <
2\varepsilon$ holds in a number of the following frames, and thus no training is required
for those frames. This way, we accelerate the tracking algorithm significantly
(Fig.~\ref{fig:temporal_modeling}). Second, we update the fusion layer in a lazy, {\emph
i.e.}, a coordinate-descent manner with a small learning rate (see
Algorithm~\ref{alg:coordinate_descent}). The learning process is thus stabilized well. In
this work, we set that $\varepsilon = 5\text{e-}3$, $r = 5\text{e-}2$ and $\hat{r} =
5\text{e-}3$. 

\section{Experiments}
\label{sec:exp}

\subsection{Benchmarks and experiment setting}

We evaluate our method on two recently proposed visual tracking benchmarks, {\it i.e.},
the CVPR2013 Visual Tracker Benchmark \cite{10.1109/CVPR.2013.312} and the VOT2013
Challenge Benchmark \cite{kristan2013visual}. These two benchmarks contain more than $60$
sequences and cover almost all the challenging scenarios such as scale changes,
illumination changes, occlusions, cluttered backgrounds and motion blur. Furthermore,
these two benchmarks evaluate tracking algorithms with different measures and criteria,
which can be used to analyze the tracker from different views. 


In the experiments on two selected benchmarks, we use the same parameter values for
DeepTrack. Most parameters of the CNN tracker are given in Sec.~\ref{sec:architecture} and
Sec.~\ref{sec:opt}. In addition, there are some motion parameters for sampling the image
patches. In this work, we only consider the displacement $\Delta_x, \Delta_y$ and the
relative scale $s$ of the object\footnote{$s = h / 32$, where $h$ is object's height}. In
a new frame, we sample $1500$ random patches in a Gaussian Distribution which centers on
the previous predicted state. The standard deviation for the three dimensions are
$\text{min}(10, 0.5\cdot h)$, $\text{min}(10, 0.5\cdot h)$ and $0.01 \cdot h$,
respectively. Note that, all parameters are fixed for all videos in both two benchmarks;
no parameter tuning is performed for any specific video sequence. We run our algorithm in
Matlab with an unoptimized code mixed with CUDA-PTX kernels for the CNN implementation.
The hardware environment includes one quad-core CPU and a NVIDIA GTX980 GPU. 

\subsection{Comparison results on the CVPR2013 benchmark}
\label{sec:comp1}

The CVPR2013 Visual Tracker Benchmark \cite{10.1109/CVPR.2013.312} contains 50 fully
annotated sequences. These sequences include many popular sequences used in the online
tracking literature over the past several years. For better evaluation and analysis of the
strength and weakness of tracking approaches, these sequences are annotated with the 11
attributes including illumination variation, scale variation, occlusion, deformation,
motion blur, fast motion, in-plane rotation, out-of-plane rotation, out-of-view,
background clutters, and low resolution. The benchmark contains the results of $29$
tracking algorithms published before the year 2013. Here, we compare our method with other
$11$ tracking methods. Among the competitors, TPGR \cite{gao2014transfer} and KCF
\cite{henriques2015tracking} are the most recently state-of-the-art visual trackers; TLD
\cite{kalal2010pn}, VTD \cite{Kwon10visualtracking}, CXT \cite{dinh2011context}, ASLA
\cite{jia2012visual}, Struck \cite{hare2011struck}, SCM \cite{zhong2012robust} are the 
top-$6$ methods as reported in the benchmark; CPF \cite{perez2002color}, IVT
\cite{Ross:2008:ILR:1345995.1346002} and MIL \cite{babenko2009visual} are classical
tracking methods which are used as comparison baselines.

The tracking results are evaluated via the following two measurements: 1) Tracking
Precision (TP), the percentage of the frames whose estimated location is within the given
distance-threshold ($\tau_d$) to the ground truth, and 2) Tracking Success Rate (TSR), the
percentage of the frames in which the overlapping score defined in
Eq.~\ref{equ:overlap_loss} between the estimated location and the ground truth is
larger than a given overlapping-threshold ($\tau_o$). Following the setting in the
recently published work \cite{gao2014transfer, henriques2015tracking}, we conduct the
experiment using the OPE (one-pass evaluation) evaluation strategy for a better
comparison to the latest methods.  

Firstly, we evaluate all algorithms using fixed thresholds, {\it i.e.}, $\tau_d = 20$,
$\tau_o = 0.6$, which is a standard setting in tracking evaluations
\cite{10.1109/CVPR.2013.312}. Results for all the involved trackers and all the video
sequences are given in Table~\ref{tab:tracking_resutls}.
According to the table, our method achieves better average performance compared with other
trackers. The performance gap between our method and the reported best result in the
literature are 6\% for the TP measure: our method achieves 83\% accuracy while the best
state-of-the-art is 77\% (TGPR method). For the TSR measure, our method is 8\% better than
the existing methods: our method gives 63\% accuracy while the best state-of-the-art is
55\% (SCM method). Furthermore, our CNN tracker have ranked as the best method for $33$
times. These numbers for TGPR, KCF, SCM and Struck are $21$, $28$, $19$ and $21$
respectively. Another observation from the Table~\ref{tab:tracking_resutls} is that,
DeepTrack rarely performs inaccurately; there are only $36$ occasions when the proposed
tracker performs significantly poorer than the best method (no less then $80\%$ of the
highest score for one sequence). 


\begin{table*}[htb]
\centering
\resizebox{1\textwidth}{!}
{

\begin{tabular}{ l | c |c |c |c |c |c |c |c |c |c |c |c }
\hline\hline
& Struck    & MIL    & VTD    & CXT    & SCM    & TLD    & ASLA    & IVT    & CPF    & KCF    & TGPR    & DeepTrack    \\

\hline\hline
\emph{tiger1} & ${0.17}/{0.13}$ & ${0.09}/{0.07}$ & ${0.12}/{0.09}$ & ${0.37}/{0.17}$ & ${0.13}/{0.11}$ & ${0.46}/{0.36}$ & ${0.23}/{0.15}$ & ${0.08}/{0.07}$ & ${0.39}/{0.24}$ & ${\color{red}0.97}/{\color{red}0.94}$ & ${0.28}/{0.22}$ & ${0.56}/{0.36}$ \\
\cline{1-13}
\emph{carDark} & ${\color{red}1.00}/{\color{red}1.00}$ & ${0.38}/{0.09}$ & ${0.74}/{0.66}$ & ${0.73}/{0.67}$ & ${\color{red}1.00}/{\color{blue}0.98}$ & ${0.64}/{0.50}$ & ${\color{red}1.00}/{\color{blue}0.99}$ & ${\color{blue}0.81}/{0.69}$ & ${0.17}/{0.02}$ & ${\color{red}1.00}/{0.44}$ & ${\color{red}1.00}/{\color{blue}0.95}$ & ${\color{red}1.00}/{\color{blue}0.97}$ \\
\cline{1-13}
\emph{girl} & ${\color{red}1.00}/{\color{red}0.90}$ & ${0.71}/{0.25}$ & ${\color{blue}0.95}/{0.41}$ & ${0.77}/{0.61}$ & ${\color{red}1.00}/{\color{blue}0.74}$ & ${\color{blue}0.92}/{0.61}$ & ${\color{red}1.00}/{\color{blue}0.78}$ & ${0.44}/{0.17}$ & ${0.74}/{0.40}$ & ${\color{blue}0.86}/{0.47}$ & ${\color{blue}0.92}/{0.69}$ & ${\color{blue}0.98}/{\color{blue}0.83}$ \\
\cline{1-13}
\emph{david} & ${0.33}/{0.19}$ & ${0.70}/{0.05}$ & ${\color{blue}0.94}/{0.38}$ & ${\color{red}1.00}/{0.48}$ & ${\color{red}1.00}/{\color{blue}0.84}$ & ${\color{red}1.00}/{\color{blue}0.83}$ & ${\color{red}1.00}/{\color{red}0.94}$ & ${\color{red}1.00}/{0.65}$ & ${0.19}/{0.02}$ & ${\color{red}1.00}/{0.26}$ & ${\color{blue}0.98}/{0.26}$ & ${\color{red}1.00}/{\color{blue}0.76}$ \\
\cline{1-13}
\emph{singer1} & ${0.64}/{0.20}$ & ${0.50}/{0.20}$ & ${\color{red}1.00}/{0.36}$ & ${\color{blue}0.97}/{0.27}$ & ${\color{red}1.00}/{\color{red}1.00}$ & ${\color{red}1.00}/{\color{blue}0.93}$ & ${\color{red}1.00}/{\color{blue}0.98}$ & ${\color{blue}0.96}/{0.35}$ & ${\color{blue}0.99}/{0.10}$ & ${\color{blue}0.81}/{0.20}$ & ${0.68}/{0.19}$ & ${\color{red}1.00}/{\color{blue}1.00}$ \\
\cline{1-13}
\emph{skating1} & ${0.47}/{0.20}$ & ${0.13}/{0.08}$ & ${\color{blue}0.90}/{\color{blue}0.43}$ & ${0.23}/{0.06}$ & ${0.77}/{0.21}$ & ${0.32}/{0.21}$ & ${0.77}/{\color{red}0.45}$ & ${0.11}/{0.05}$ & ${0.23}/{0.17}$ & ${\color{red}1.00}/{0.23}$ & ${\color{blue}0.81}/{0.25}$ & ${\color{red}1.00}/{\color{red}0.45}$ \\
\cline{1-13}
\emph{deer} & ${\color{red}1.00}/{\color{blue}0.94}$ & ${0.13}/{0.07}$ & ${0.04}/{0.03}$ & ${\color{red}1.00}/{\color{blue}0.87}$ & ${0.03}/{0.03}$ & ${0.73}/{0.73}$ & ${0.03}/{0.03}$ & ${0.03}/{0.03}$ & ${0.04}/{0.03}$ & ${\color{blue}0.82}/{0.76}$ & ${\color{blue}0.86}/{0.79}$ & ${\color{red}1.00}/{\color{red}0.99}$ \\
\cline{1-13}
\emph{singer2} & ${0.04}/{0.03}$ & ${0.40}/{0.27}$ & ${0.45}/{0.43}$ & ${0.06}/{0.04}$ & ${0.11}/{0.13}$ & ${0.07}/{0.05}$ & ${0.04}/{0.03}$ & ${0.04}/{0.04}$ & ${0.12}/{0.09}$ & ${\color{blue}0.95}/{\color{blue}0.89}$ & ${\color{red}0.97}/{\color{red}0.91}$ & ${0.57}/{0.34}$ \\
\cline{1-13}
\emph{car4} & ${\color{blue}0.99}/{0.26}$ & ${0.35}/{0.23}$ & ${0.36}/{0.32}$ & ${0.38}/{0.27}$ & ${\color{blue}0.97}/{\color{blue}0.93}$ & ${\color{blue}0.87}/{0.63}$ & ${\color{red}1.00}/{\color{blue}0.95}$ & ${\color{red}1.00}/{\color{red}1.00}$ & ${0.14}/{0.01}$ & ${\color{blue}0.95}/{0.24}$ & ${\color{red}1.00}/{0.28}$ & ${\color{red}1.00}/{\color{red}1.00}$ \\
\cline{1-13}
\emph{tiger2} & ${\color{blue}0.63}/{\color{blue}0.42}$ & ${0.41}/{0.23}$ & ${0.16}/{0.08}$ & ${0.34}/{0.16}$ & ${0.11}/{0.05}$ & ${0.39}/{0.04}$ & ${0.14}/{0.11}$ & ${0.08}/{0.05}$ & ${0.11}/{0.04}$ & ${0.36}/{0.28}$ & ${\color{red}0.72}/{\color{red}0.47}$ & ${0.49}/{0.32}$ \\
\cline{1-13}
\emph{dudek} & ${\color{red}0.90}/{\color{blue}0.81}$ & ${0.69}/{0.76}$ & ${\color{blue}0.88}/{\color{red}0.96}$ & ${\color{blue}0.82}/{\color{blue}0.87}$ & ${\color{blue}0.88}/{\color{blue}0.86}$ & ${0.60}/{0.63}$ & ${\color{blue}0.75}/{0.74}$ & ${\color{blue}0.89}/{\color{blue}0.88}$ & ${0.57}/{0.58}$ & ${\color{blue}0.88}/{\color{blue}0.82}$ & ${\color{blue}0.75}/{0.71}$ & ${\color{blue}0.73}/{\color{blue}0.81}$ \\
\cline{1-13}
\emph{sylvester} & ${\color{blue}0.99}/{\color{blue}0.85}$ & ${0.65}/{0.46}$ & ${\color{blue}0.82}/{0.74}$ & ${\color{blue}0.85}/{0.56}$ & ${\color{blue}0.95}/{\color{blue}0.77}$ & ${\color{blue}0.95}/{\color{blue}0.80}$ & ${\color{blue}0.82}/{0.65}$ & ${0.68}/{0.63}$ & ${\color{blue}0.86}/{0.52}$ & ${\color{blue}0.84}/{0.73}$ & ${\color{blue}0.96}/{\color{red}0.93}$ & ${\color{red}1.00}/{\color{blue}0.92}$ \\
\cline{1-13}
\emph{jumping} & ${\color{red}1.00}/{0.50}$ & ${\color{blue}1.00}/{0.33}$ & ${0.21}/{0.08}$ & ${\color{blue}1.00}/{0.25}$ & ${0.15}/{0.11}$ & ${\color{red}1.00}/{0.70}$ & ${0.45}/{0.15}$ & ${0.21}/{0.08}$ & ${0.16}/{0.09}$ & ${0.34}/{0.26}$ & ${\color{blue}0.95}/{0.50}$ & ${\color{red}1.00}/{\color{red}0.93}$ \\
\cline{1-13}
\emph{david2} & ${\color{red}1.00}/{\color{red}1.00}$ & ${\color{blue}0.98}/{0.24}$ & ${\color{red}1.00}/{\color{blue}0.88}$ & ${\color{red}1.00}/{\color{red}1.00}$ & ${\color{red}1.00}/{0.80}$ & ${\color{red}1.00}/{0.70}$ & ${\color{red}1.00}/{\color{blue}0.95}$ & ${\color{red}1.00}/{0.74}$ & ${\color{red}1.00}/{0.25}$ & ${\color{red}1.00}/{\color{red}1.00}$ & ${\color{red}1.00}/{\color{blue}0.97}$ & ${\color{red}1.00}/{\color{blue}0.87}$ \\
\cline{1-13}
\emph{shaking} & ${0.19}/{0.04}$ & ${0.28}/{0.18}$ & ${\color{blue}0.93}/{\color{red}0.83}$ & ${0.13}/{0.04}$ & ${\color{blue}0.81}/{\color{blue}0.69}$ & ${0.41}/{0.31}$ & ${0.48}/{0.17}$ & ${0.01}/{0.01}$ & ${0.17}/{0.07}$ & ${0.02}/{0.01}$ & ${\color{red}0.97}/{\color{blue}0.70}$ & ${\color{blue}0.95}/{\color{blue}0.68}$ \\
\cline{1-13}
\emph{trellis} & ${\color{blue}0.88}/{0.72}$ & ${0.23}/{0.16}$ & ${0.50}/{0.44}$ & ${\color{blue}0.97}/{0.69}$ & ${\color{blue}0.87}/{\color{blue}0.84}$ & ${0.53}/{0.45}$ & ${\color{blue}0.86}/{\color{blue}0.85}$ & ${0.33}/{0.26}$ & ${0.30}/{0.14}$ & ${\color{red}1.00}/{0.74}$ & ${\color{blue}0.98}/{0.68}$ & ${\color{red}1.00}/{\color{red}0.96}$ \\
\cline{1-13}
\emph{woman} & ${\color{red}1.00}/{\color{blue}0.89}$ & ${0.21}/{0.18}$ & ${0.20}/{0.16}$ & ${0.37}/{0.15}$ & ${\color{blue}0.94}/{0.69}$ & ${0.19}/{0.15}$ & ${0.20}/{0.17}$ & ${0.20}/{0.17}$ & ${0.20}/{0.05}$ & ${\color{blue}0.94}/{\color{red}0.90}$ & ${\color{blue}0.97}/{\color{blue}0.87}$ & ${\color{blue}0.98}/{0.24}$ \\
\cline{1-13}
\emph{fish} & ${\color{red}1.00}/{\color{red}1.00}$ & ${0.39}/{0.28}$ & ${0.65}/{0.57}$ & ${\color{red}1.00}/{\color{red}1.00}$ & ${\color{blue}0.86}/{\color{blue}0.85}$ & ${\color{red}1.00}/{\color{blue}0.96}$ & ${\color{red}1.00}/{\color{red}1.00}$ & ${\color{red}1.00}/{\color{red}1.00}$ & ${0.11}/{0.08}$ & ${\color{red}1.00}/{\color{red}1.00}$ & ${\color{blue}0.97}/{\color{blue}0.97}$ & ${\color{red}1.00}/{\color{red}1.00}$ \\
\cline{1-13}
\emph{matrix} & ${0.12}/{0.12}$ & ${0.18}/{0.10}$ & ${0.22}/{0.03}$ & ${0.06}/{0.01}$ & ${0.35}/{0.24}$ & ${0.16}/{0.03}$ & ${0.05}/{0.01}$ & ${0.02}/{0.02}$ & ${0.09}/{0.02}$ & ${0.17}/{0.11}$ & ${0.39}/{0.26}$ & ${\color{red}0.72}/{\color{red}0.43}$ \\
\cline{1-13}
\emph{ironman} & ${0.11}/{0.02}$ & ${0.11}/{0.02}$ & ${\color{blue}0.17}/{\color{blue}0.12}$ & ${0.04}/{0.03}$ & ${0.16}/{0.09}$ & ${0.12}/{0.04}$ & ${0.13}/{0.08}$ & ${0.05}/{0.05}$ & ${0.05}/{0.04}$ & ${\color{red}0.22}/{0.10}$ & ${\color{red}0.22}/{\color{red}0.13}$ & ${0.08}/{0.05}$ \\
\cline{1-13}
\emph{mhyang} & ${\color{red}1.00}/{\color{blue}0.97}$ & ${0.46}/{0.25}$ & ${\color{red}1.00}/{0.77}$ & ${\color{red}1.00}/{\color{red}1.00}$ & ${\color{red}1.00}/{\color{blue}0.96}$ & ${\color{blue}0.98}/{0.52}$ & ${\color{red}1.00}/{\color{blue}1.00}$ & ${\color{red}1.00}/{\color{blue}1.00}$ & ${0.79}/{0.08}$ & ${\color{red}1.00}/{\color{blue}0.93}$ & ${\color{blue}0.95}/{\color{blue}0.88}$ & ${\color{red}1.00}/{\color{blue}0.96}$ \\
\cline{1-13}
\emph{liquor} & ${0.39}/{0.40}$ & ${0.20}/{0.20}$ & ${0.52}/{0.52}$ & ${0.21}/{0.21}$ & ${0.28}/{0.29}$ & ${0.59}/{0.54}$ & ${0.23}/{0.23}$ & ${0.21}/{0.21}$ & ${0.52}/{0.53}$ & ${\color{red}0.98}/{\color{red}0.97}$ & ${0.27}/{0.27}$ & ${\color{blue}0.91}/{\color{blue}0.89}$ \\
\cline{1-13}
\emph{motorRolling} & ${0.09}/{0.09}$ & ${0.04}/{0.06}$ & ${0.05}/{0.05}$ & ${0.04}/{0.02}$ & ${0.04}/{0.05}$ & ${0.12}/{0.10}$ & ${0.06}/{0.07}$ & ${0.03}/{0.04}$ & ${0.06}/{0.04}$ & ${0.05}/{0.05}$ & ${0.09}/{0.10}$ & ${\color{red}0.80}/{\color{red}0.43}$ \\
\cline{1-13}
\emph{coke} & ${\color{red}0.95}/{\color{red}0.87}$ & ${0.15}/{0.08}$ & ${0.15}/{0.11}$ & ${0.65}/{0.15}$ & ${0.43}/{0.24}$ & ${0.68}/{0.09}$ & ${0.16}/{0.10}$ & ${0.13}/{0.13}$ & ${0.39}/{0.03}$ & ${\color{blue}0.84}/{0.41}$ & ${\color{blue}0.95}/{0.63}$ & ${\color{blue}0.91}/{0.18}$ \\
\cline{1-13}
\emph{soccer} & ${0.25}/{0.15}$ & ${0.19}/{0.14}$ & ${0.45}/{0.18}$ & ${0.23}/{0.12}$ & ${0.27}/{0.16}$ & ${0.11}/{0.11}$ & ${0.12}/{0.11}$ & ${0.17}/{0.14}$ & ${0.26}/{0.16}$ & ${\color{red}0.79}/{\color{red}0.35}$ & ${0.16}/{0.13}$ & ${0.30}/{0.16}$ \\
\cline{1-13}
\emph{boy} & ${\color{red}1.00}/{\color{blue}0.93}$ & ${\color{blue}0.85}/{0.29}$ & ${\color{blue}0.97}/{0.61}$ & ${\color{blue}0.94}/{0.42}$ & ${0.44}/{0.44}$ & ${\color{red}1.00}/{0.74}$ & ${0.44}/{0.44}$ & ${0.33}/{0.31}$ & ${\color{blue}1.00}/{\color{blue}0.82}$ & ${\color{red}1.00}/{\color{red}0.96}$ & ${\color{blue}0.99}/{\color{blue}0.91}$ & ${\color{blue}1.00}/{\color{blue}0.93}$ \\
\cline{1-13}
\emph{basketball} & ${0.12}/{0.09}$ & ${0.28}/{0.20}$ & ${\color{red}1.00}/{\color{red}0.85}$ & ${0.04}/{0.02}$ & ${0.66}/{0.53}$ & ${0.03}/{0.02}$ & ${0.60}/{0.26}$ & ${0.50}/{0.08}$ & ${0.74}/{0.54}$ & ${\color{blue}0.92}/{\color{blue}0.71}$ & ${\color{blue}0.99}/{\color{blue}0.69}$ & ${\color{blue}0.82}/{0.39}$ \\
\cline{1-13}
\emph{lemming} & ${0.63}/{0.49}$ & ${\color{blue}0.82}/{\color{red}0.68}$ & ${0.51}/{0.42}$ & ${\color{blue}0.73}/{0.38}$ & ${0.17}/{0.16}$ & ${\color{blue}0.86}/{0.43}$ & ${0.17}/{0.17}$ & ${0.17}/{0.17}$ & ${\color{red}0.88}/{0.40}$ & ${0.49}/{0.30}$ & ${0.35}/{0.26}$ & ${0.28}/{0.26}$ \\
\cline{1-13}
\emph{bolt} & ${0.02}/{0.01}$ & ${0.01}/{0.01}$ & ${0.31}/{0.14}$ & ${0.03}/{0.01}$ & ${0.03}/{0.01}$ & ${0.31}/{0.08}$ & ${0.02}/{0.01}$ & ${0.01}/{0.01}$ & ${\color{blue}0.91}/{0.15}$ & ${\color{blue}0.99}/{\color{blue}0.75}$ & ${0.02}/{0.01}$ & ${\color{red}0.99}/{\color{red}0.78}$ \\
\cline{1-13}
\emph{crossing} & ${\color{red}1.00}/{0.72}$ & ${\color{red}1.00}/{\color{blue}0.83}$ & ${0.44}/{0.36}$ & ${0.62}/{0.32}$ & ${\color{red}1.00}/{\color{red}0.99}$ & ${0.62}/{0.41}$ & ${\color{red}1.00}/{\color{red}0.99}$ & ${\color{red}1.00}/{0.23}$ & ${\color{blue}0.89}/{0.38}$ & ${\color{red}1.00}/{0.78}$ & ${\color{red}1.00}/{\color{blue}0.81}$ & ${\color{blue}0.94}/{0.56}$ \\
\cline{1-13}
\emph{couple} & ${0.74}/{0.51}$ & ${0.68}/{0.61}$ & ${0.11}/{0.06}$ & ${0.64}/{0.52}$ & ${0.11}/{0.11}$ & ${\color{red}1.00}/{\color{red}0.98}$ & ${0.09}/{0.09}$ & ${0.09}/{0.09}$ & ${\color{blue}0.87}/{0.58}$ & ${0.26}/{0.24}$ & ${0.60}/{0.35}$ & ${\color{blue}0.99}/{0.63}$ \\
\cline{1-13}
\emph{david3} & ${0.34}/{0.34}$ & ${0.74}/{0.60}$ & ${0.56}/{0.44}$ & ${0.15}/{0.10}$ & ${0.50}/{0.47}$ & ${0.11}/{0.10}$ & ${0.55}/{0.49}$ & ${0.75}/{0.41}$ & ${0.57}/{0.33}$ & ${\color{red}1.00}/{\color{red}0.96}$ & ${\color{blue}1.00}/{0.69}$ & ${\color{blue}1.00}/{\color{blue}0.93}$ \\
\cline{1-13}
\emph{carScale} & ${0.65}/{0.37}$ & ${0.63}/{0.35}$ & ${0.55}/{0.42}$ & ${\color{blue}0.74}/{\color{red}0.74}$ & ${0.65}/{\color{blue}0.64}$ & ${\color{red}0.85}/{0.29}$ & ${\color{blue}0.74}/{\color{blue}0.65}$ & ${\color{blue}0.78}/{\color{blue}0.67}$ & ${0.67}/{0.32}$ & ${\color{blue}0.81}/{0.35}$ & ${\color{blue}0.79}/{0.37}$ & ${0.67}/{0.56}$ \\
\cline{1-13}
\emph{doll} & ${\color{blue}0.92}/{0.34}$ & ${0.73}/{0.20}$ & ${\color{blue}0.97}/{0.73}$ & ${\color{red}0.99}/{\color{blue}0.87}$ & ${\color{blue}0.98}/{\color{red}0.97}$ & ${\color{blue}0.98}/{0.39}$ & ${\color{blue}0.92}/{\color{blue}0.91}$ & ${0.76}/{0.27}$ & ${\color{blue}0.94}/{\color{blue}0.84}$ & ${\color{blue}0.97}/{0.33}$ & ${\color{blue}0.94}/{0.40}$ & ${\color{blue}0.96}/{\color{blue}0.86}$ \\
\cline{1-13}
\emph{skiing} & ${0.04}/{0.04}$ & ${0.07}/{0.06}$ & ${\color{blue}0.14}/{0.01}$ & ${\color{red}0.15}/{0.06}$ & ${\color{blue}0.14}/{0.06}$ & ${\color{blue}0.12}/{0.05}$ & ${\color{blue}0.14}/{\color{red}0.11}$ & ${0.11}/{0.09}$ & ${0.06}/{0.01}$ & ${0.07}/{0.05}$ & ${\color{blue}0.12}/{\color{blue}0.10}$ & ${0.09}/{0.06}$ \\
\cline{1-13}
\emph{football} & ${0.75}/{0.57}$ & ${0.79}/{\color{blue}0.67}$ & ${0.80}/{\color{blue}0.65}$ & ${0.80}/{0.57}$ & ${0.77}/{0.42}$ & ${\color{blue}0.80}/{0.28}$ & ${0.73}/{\color{blue}0.62}$ & ${0.79}/{\color{blue}0.61}$ & ${\color{blue}0.97}/{\color{blue}0.60}$ & ${0.80}/{0.57}$ & ${\color{red}1.00}/{\color{red}0.75}$ & ${0.79}/{0.52}$ \\
\cline{1-13}
\emph{football1} & ${\color{red}1.00}/{0.72}$ & ${\color{red}1.00}/{0.55}$ & ${\color{blue}0.99}/{0.51}$ & ${\color{red}1.00}/{\color{red}0.96}$ & ${0.57}/{0.34}$ & ${0.55}/{0.34}$ & ${0.80}/{0.39}$ & ${\color{blue}0.81}/{0.49}$ & ${\color{red}1.00}/{0.58}$ & ${\color{blue}0.96}/{\color{blue}0.80}$ & ${\color{blue}0.99}/{0.41}$ & ${\color{red}1.00}/{0.38}$ \\
\cline{1-13}
\emph{freeman1} & ${\color{blue}0.80}/{0.16}$ & ${\color{blue}0.94}/{0.12}$ & ${\color{blue}0.95}/{0.13}$ & ${0.73}/{0.18}$ & ${\color{blue}0.98}/{\color{red}0.54}$ & ${0.54}/{0.18}$ & ${0.39}/{0.20}$ & ${\color{blue}0.81}/{0.26}$ & ${0.76}/{0.18}$ & ${0.39}/{0.13}$ & ${\color{blue}0.93}/{0.21}$ & ${\color{red}1.00}/{0.35}$ \\
\cline{1-13}
\emph{freeman3} & ${0.79}/{0.12}$ & ${0.05}/{0.00}$ & ${0.72}/{0.22}$ & ${\color{red}1.00}/{\color{blue}0.89}$ & ${\color{red}1.00}/{\color{blue}0.88}$ & ${0.77}/{0.42}$ & ${\color{red}1.00}/{\color{red}0.90}$ & ${0.76}/{0.33}$ & ${0.17}/{0.14}$ & ${\color{blue}0.91}/{0.21}$ & ${0.77}/{0.15}$ & ${\color{blue}0.97}/{0.67}$ \\
\cline{1-13}
\emph{freeman4} & ${0.37}/{0.15}$ & ${0.20}/{0.02}$ & ${0.37}/{0.08}$ & ${0.43}/{0.17}$ & ${0.51}/{0.18}$ & ${0.41}/{\color{red}0.24}$ & ${0.22}/{0.16}$ & ${0.35}/{0.17}$ & ${0.12}/{0.02}$ & ${0.53}/{0.12}$ & ${\color{blue}0.58}/{\color{blue}0.21}$ & ${\color{red}0.71}/{\color{blue}0.22}$ \\
\cline{1-13}
\emph{subway} & ${\color{blue}0.98}/{0.63}$ & ${\color{blue}0.99}/{0.68}$ & ${0.23}/{0.18}$ & ${0.26}/{0.20}$ & ${\color{red}1.00}/{\color{blue}0.90}$ & ${0.25}/{0.22}$ & ${0.23}/{0.21}$ & ${0.22}/{0.19}$ & ${0.22}/{0.10}$ & ${\color{red}1.00}/{\color{blue}0.94}$ & ${\color{red}1.00}/{\color{red}0.99}$ & ${\color{red}1.00}/{0.79}$ \\
\cline{1-13}
\emph{suv} & ${0.57}/{0.57}$ & ${0.12}/{0.12}$ & ${0.52}/{0.47}$ & ${\color{blue}0.91}/{\color{blue}0.90}$ & ${\color{blue}0.98}/{\color{blue}0.80}$ & ${\color{blue}0.91}/{0.70}$ & ${0.57}/{0.55}$ & ${0.45}/{0.44}$ & ${0.78}/{0.63}$ & ${\color{red}0.98}/{\color{red}0.98}$ & ${0.66}/{0.66}$ & ${0.52}/{0.52}$ \\
\cline{1-13}
\emph{walking} & ${\color{red}1.00}/{0.42}$ & ${\color{red}1.00}/{0.37}$ & ${\color{red}1.00}/{0.55}$ & ${0.24}/{0.22}$ & ${\color{red}1.00}/{\color{blue}0.86}$ & ${\color{blue}0.96}/{0.30}$ & ${\color{red}1.00}/{\color{red}0.99}$ & ${\color{red}1.00}/{\color{blue}0.98}$ & ${\color{red}1.00}/{0.65}$ & ${\color{red}1.00}/{0.34}$ & ${\color{red}1.00}/{0.41}$ & ${\color{red}1.00}/{\color{blue}0.94}$ \\
\cline{1-13}
\emph{walking2} & ${\color{blue}0.98}/{0.32}$ & ${0.41}/{0.31}$ & ${0.41}/{0.39}$ & ${0.41}/{0.39}$ & ${\color{red}1.00}/{\color{blue}0.99}$ & ${0.43}/{0.29}$ & ${0.40}/{0.40}$ & ${\color{red}1.00}/{\color{red}0.99}$ & ${0.36}/{0.35}$ & ${0.44}/{0.30}$ & ${\color{blue}0.99}/{0.31}$ & ${0.61}/{0.38}$ \\
\cline{1-13}
\emph{mountainBike} & ${\color{blue}0.92}/{0.67}$ & ${0.67}/{0.41}$ & ${\color{blue}1.00}/{\color{blue}0.81}$ & ${0.28}/{0.28}$ & ${\color{blue}0.97}/{0.72}$ & ${0.26}/{0.21}$ & ${\color{blue}0.90}/{\color{blue}0.82}$ & ${\color{blue}1.00}/{\color{blue}0.84}$ & ${0.15}/{0.06}$ & ${\color{red}1.00}/{\color{blue}0.88}$ & ${\color{red}1.00}/{\color{blue}0.87}$ & ${\color{red}1.00}/{\color{red}0.91}$ \\
\cline{1-13}
\emph{faceocc1} & ${0.58}/{\color{blue}0.95}$ & ${0.22}/{0.46}$ & ${0.53}/{0.72}$ & ${0.34}/{0.57}$ & ${\color{red}0.93}/{\color{red}1.00}$ & ${0.20}/{0.65}$ & ${0.18}/{0.25}$ & ${0.64}/{\color{blue}0.87}$ & ${0.32}/{0.41}$ & ${0.73}/{\color{blue}0.99}$ & ${0.66}/{0.80}$ & ${0.33}/{0.42}$ \\
\cline{1-13}
\emph{jogging-1} & ${0.24}/{0.22}$ & ${0.23}/{0.21}$ & ${0.23}/{0.18}$ & ${\color{blue}0.96}/{\color{blue}0.95}$ & ${0.23}/{0.21}$ & ${\color{blue}0.97}/{\color{blue}0.95}$ & ${0.23}/{0.22}$ & ${0.22}/{0.22}$ & ${0.54}/{0.23}$ & ${0.23}/{0.22}$ & ${\color{red}0.99}/{\color{red}0.96}$ & ${\color{blue}0.97}/{\color{blue}0.94}$ \\
\cline{1-13}
\emph{jogging-2} & ${0.25}/{0.22}$ & ${0.19}/{0.16}$ & ${0.19}/{0.16}$ & ${0.16}/{0.15}$ & ${\color{red}1.00}/{\color{red}0.98}$ & ${\color{blue}0.86}/{\color{blue}0.83}$ & ${0.18}/{0.17}$ & ${0.20}/{0.19}$ & ${\color{blue}0.84}/{0.72}$ & ${0.16}/{0.15}$ & ${\color{blue}1.00}/{\color{blue}0.95}$ & ${\color{blue}0.99}/{0.30}$ \\
\cline{1-13}
\emph{dog1} & ${\color{blue}1.00}/{0.51}$ & ${\color{blue}0.92}/{0.45}$ & ${\color{blue}0.83}/{0.61}$ & ${\color{red}1.00}/{\color{blue}0.95}$ & ${\color{blue}0.98}/{0.76}$ & ${\color{red}1.00}/{0.61}$ & ${\color{blue}1.00}/{\color{blue}0.87}$ & ${\color{blue}0.98}/{\color{blue}0.80}$ & ${\color{blue}0.91}/{\color{blue}0.90}$ & ${\color{red}1.00}/{0.51}$ & ${\color{red}1.00}/{0.52}$ & ${\color{red}1.00}/{\color{red}0.95}$ \\
\cline{1-13}
\emph{fleetface} & ${\color{blue}0.64}/{0.51}$ & ${0.36}/{0.32}$ & ${\color{red}0.66}/{\color{red}0.68}$ & ${\color{blue}0.57}/{\color{blue}0.60}$ & ${\color{blue}0.53}/{\color{blue}0.58}$ & ${0.51}/{0.41}$ & ${0.30}/{0.32}$ & ${0.26}/{0.24}$ & ${0.16}/{0.21}$ & ${0.46}/{0.47}$ & ${0.45}/{0.47}$ & ${0.51}/{\color{blue}0.60}$ \\
\cline{1-13}
\emph{faceocc2} & ${\color{red}1.00}/{\color{red}0.97}$ & ${0.74}/{0.62}$ & ${\color{blue}0.98}/{\color{blue}0.84}$ & ${\color{red}1.00}/{\color{blue}0.90}$ & ${\color{blue}0.86}/{0.74}$ & ${\color{blue}0.86}/{0.51}$ & ${0.79}/{0.61}$ & ${\color{blue}0.99}/{0.77}$ & ${0.40}/{0.29}$ & ${\color{blue}0.97}/{\color{blue}0.79}$ & ${0.47}/{0.45}$ & ${\color{blue}1.00}/{0.71}$ \\
\cline{1-13}
{\bf Overall} & ${\bf 0.66}/{\bf 0.48}$ & ${\bf 0.47}/{\bf 0.28}$ & ${\bf 0.58}/{\bf 0.41}$ & ${\bf 0.58}/{\bf 0.43}$ & ${\bf 0.65}/{\bf 0.55}$ & ${\bf 0.61}/{\bf 0.42}$ & ${\bf 0.53}/{\bf 0.46}$ & ${\bf 0.50}/{\bf 0.38}$ & ${\bf 0.49}/{\bf 0.28}$ & ${\bf 0.74}/{\bf 0.53}$ & ${\bf 0.77}/{\bf 0.54}$ & ${\color{red}\bf 0.83}/{\color{red}\bf 0.63}$ \\
\cline{1-13}
{\bf No. Best} & $\bf 21$ & $\bf 4$ & $\bf 10$ & $\bf 16$ & $\bf 19$ & $\bf 11$ & $\bf 18$ & $\bf 11$ & $\bf 4$ & $\bf 28$ & $\bf 21$ & ${\color{red}\bf 33}$ \\
\cline{1-13}
{\bf No. Bad} & $\bf 62$ & $\bf 89$ & $\bf 71$ & $\bf 66$ & $\bf 51$ & $\bf 72$ & $\bf 64$ & $\bf 74$ & $\bf 84$ & $\bf 48$ & $\bf 45$ & ${\color{red}\bf 36}$ \\

\hline\hline
\end{tabular}
}
\caption
{
  The tracking scores of DeepTrack and other visual trackers on the CVPR2013 benchmark.
  The reported results are shown in the order of ``TP/TSR''. The top scores are shown in
  red for each row. A score is shown in blue if it is higher than $80\%$ of the highest
  value in that row.  ``No. Best'' row shows the number of best scores for each tracking
  algorithm while ``No. Bad'' row shows the number of low scores, {\emph i.e.}, the scores
  lower than $80\%$ of the maximum one in the corresponding row.
}
\label{tab:tracking_resutls}
\vspace{-0.1in}
\end{table*}

\begin{figure*}[ht!]
  \centering
  \subfigure{\includegraphics[width=0.48\textwidth]{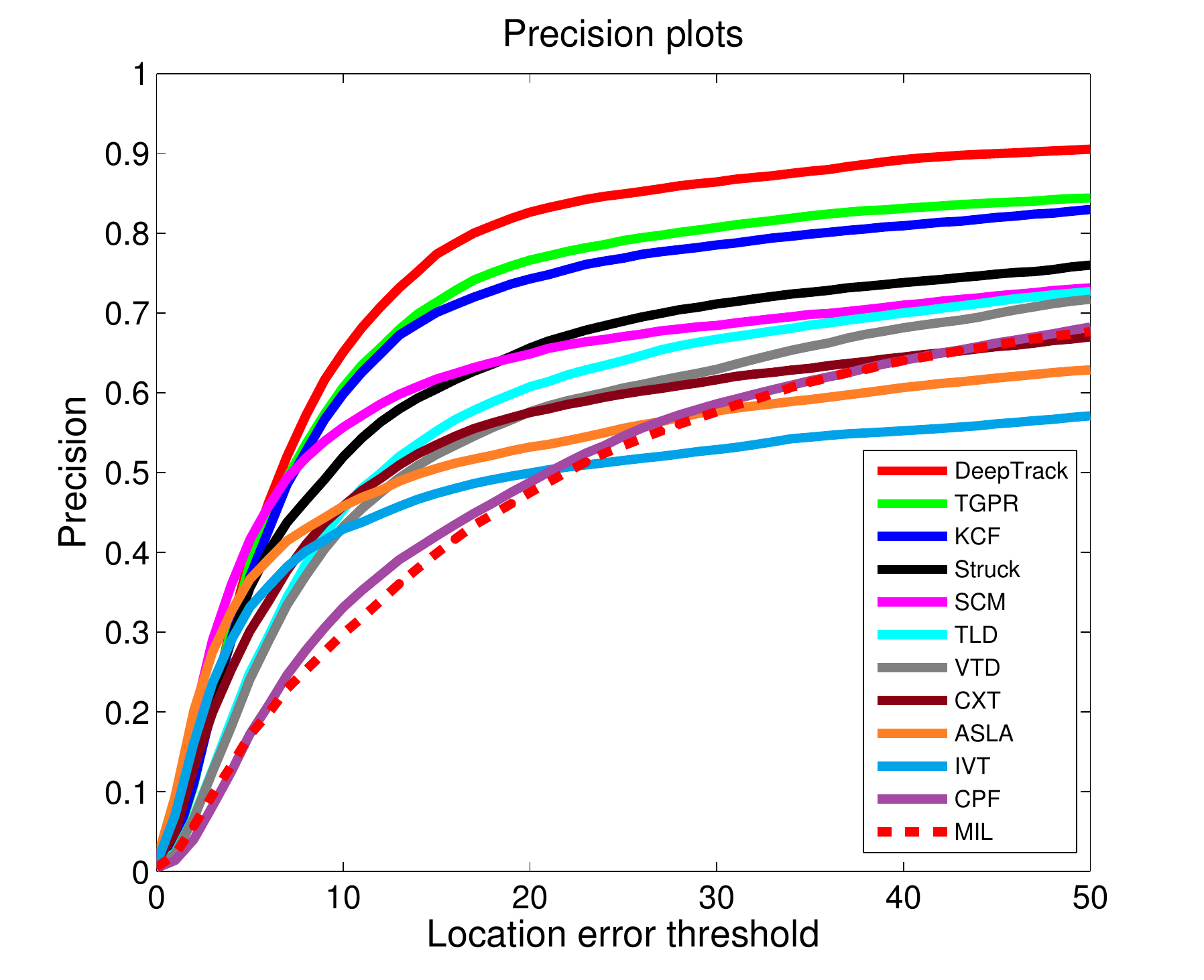}}
  \subfigure{\includegraphics[width=0.48\textwidth]{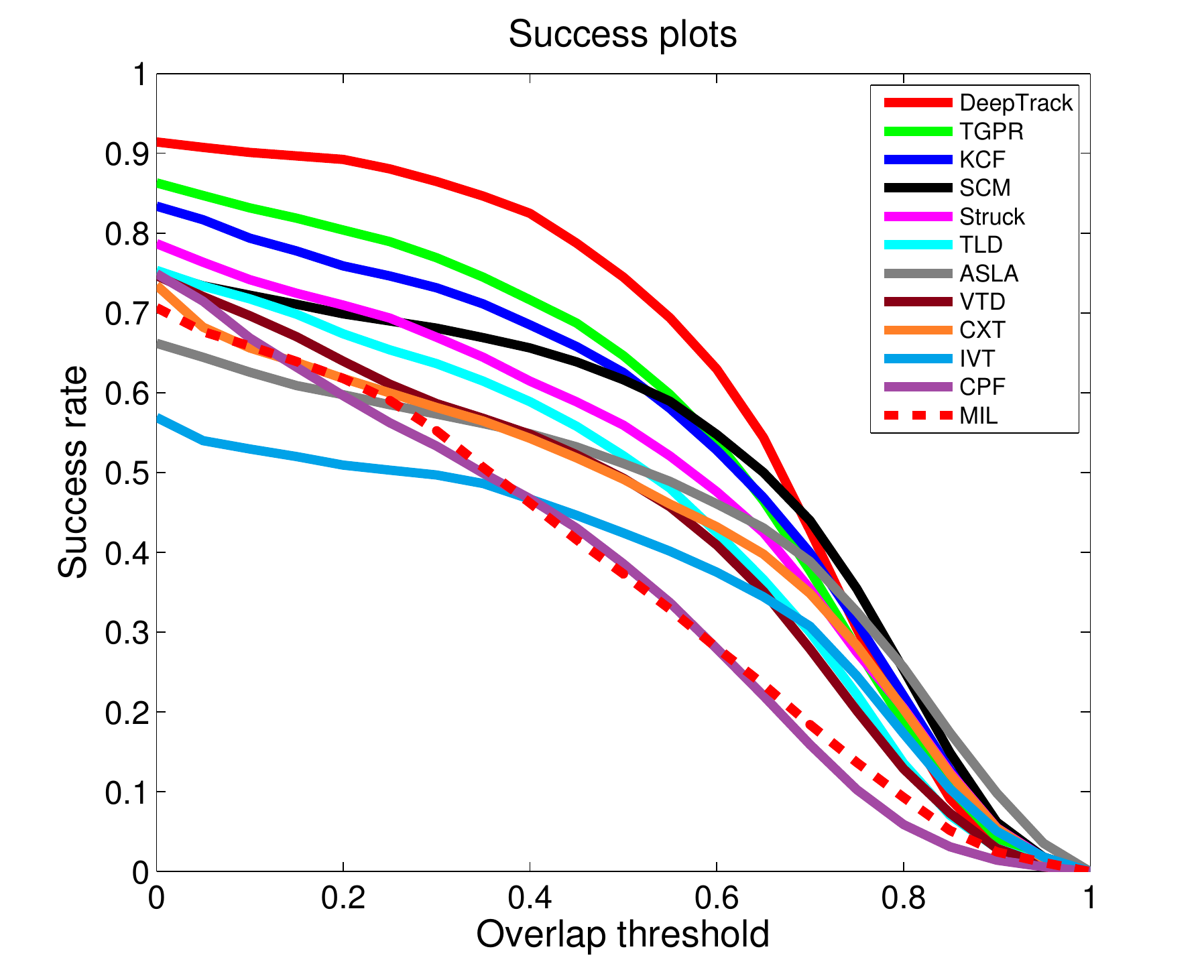}} \\
  \caption
  {
    The Precision Plot (left) and the Success Plot (right) of the tracking results on the
    CVPR2013 benchmark. Note that the color of one curve is determined by the rank of the
    corresponding trackers, not their names. 
  }
\label{fig:pre_suc}%
\end{figure*}

In fact, the superiority of our method becomes more clear when the tracking result are
evaluated using different measurement criteria (different $\tau_d$, $\tau_o$). In
specific, for TP, we evaluate the trackers with the thresholds $\tau_d = 1, 2, \cdots, 50$
while for TSR, we use the thresholds $\tau_o = 0$ to $1$ at the step of $0.05$.
Accordingly we generate the precision curves and the success-rate curves for each tracking
method, which is shown in Fig.~\ref{fig:pre_suc}.

From the score plots we can see that, overall the CNN tracker ranks the first (red curves)
for both TP and TSR evaluations. The proposed DeepTrack method outperform all the other
trackers when $\tau_o < 0.68$ and $\tau_d > 10$. When the evaluation threshold is
reasonably loose, ({\it i.e.}, $\tau_o < 0.45$ and $\tau_d > 20$), our algorithm is very robust
with both the accuracies higher than $80\%$. 
Having mentioned that when the overlap thresholds are tight (\textit{e.g.} $\tau_o > 0.75$
or $\tau_d < 5$), our tracker has similar response to rest of the trackers we tested. 

In many applications, it is more important to not to loose the target object than very
accurately locate its bounding box. As visible, our tracker rarely looses the object. It
achieves the accuracies around $90\%$ when $\tau_o < 0.3$ and $\tau_d > 30$.

\begin{figure*}[ht!]
  \centering
  \subfigure{\includegraphics[width=0.24\textwidth]{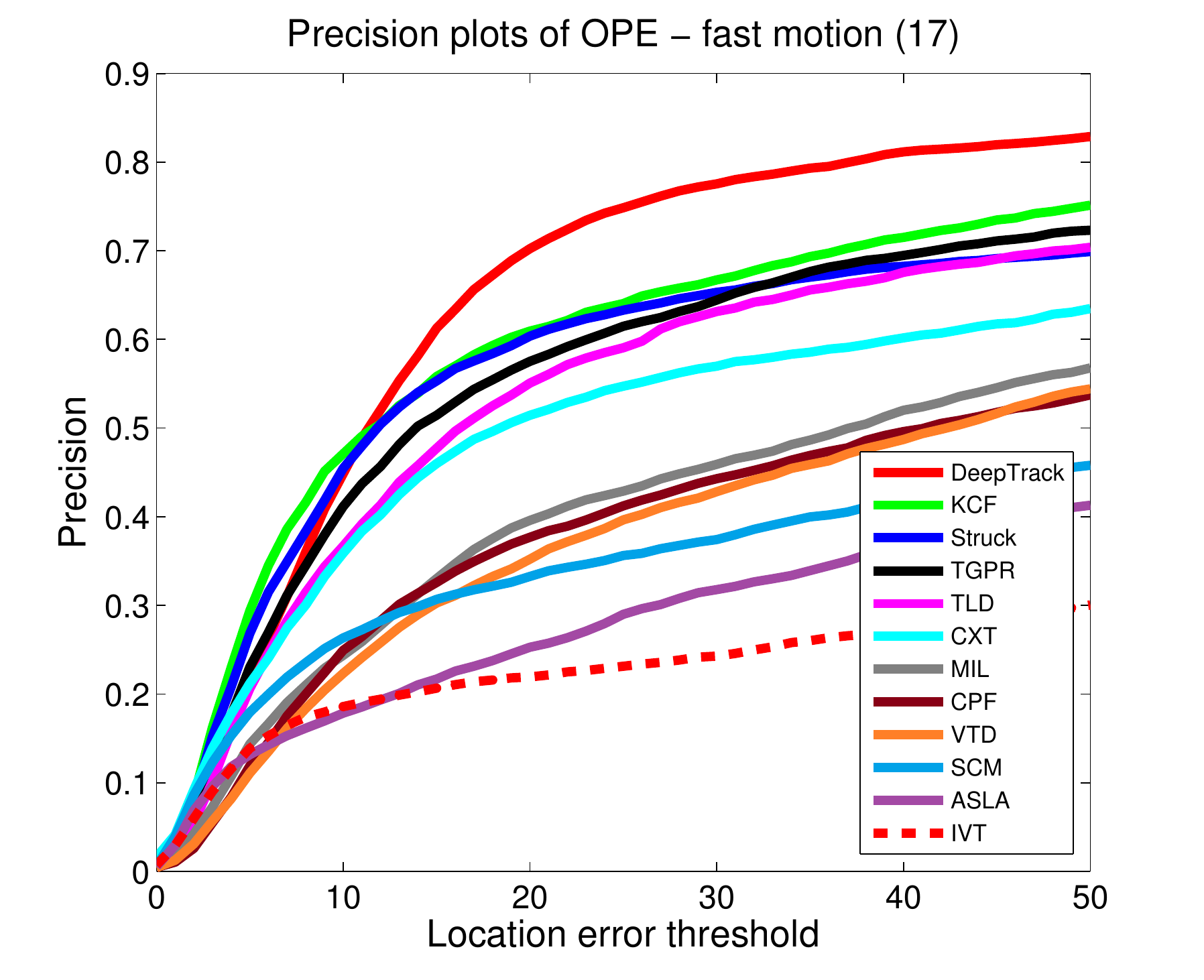}}
  \subfigure{\includegraphics[width=0.24\textwidth]{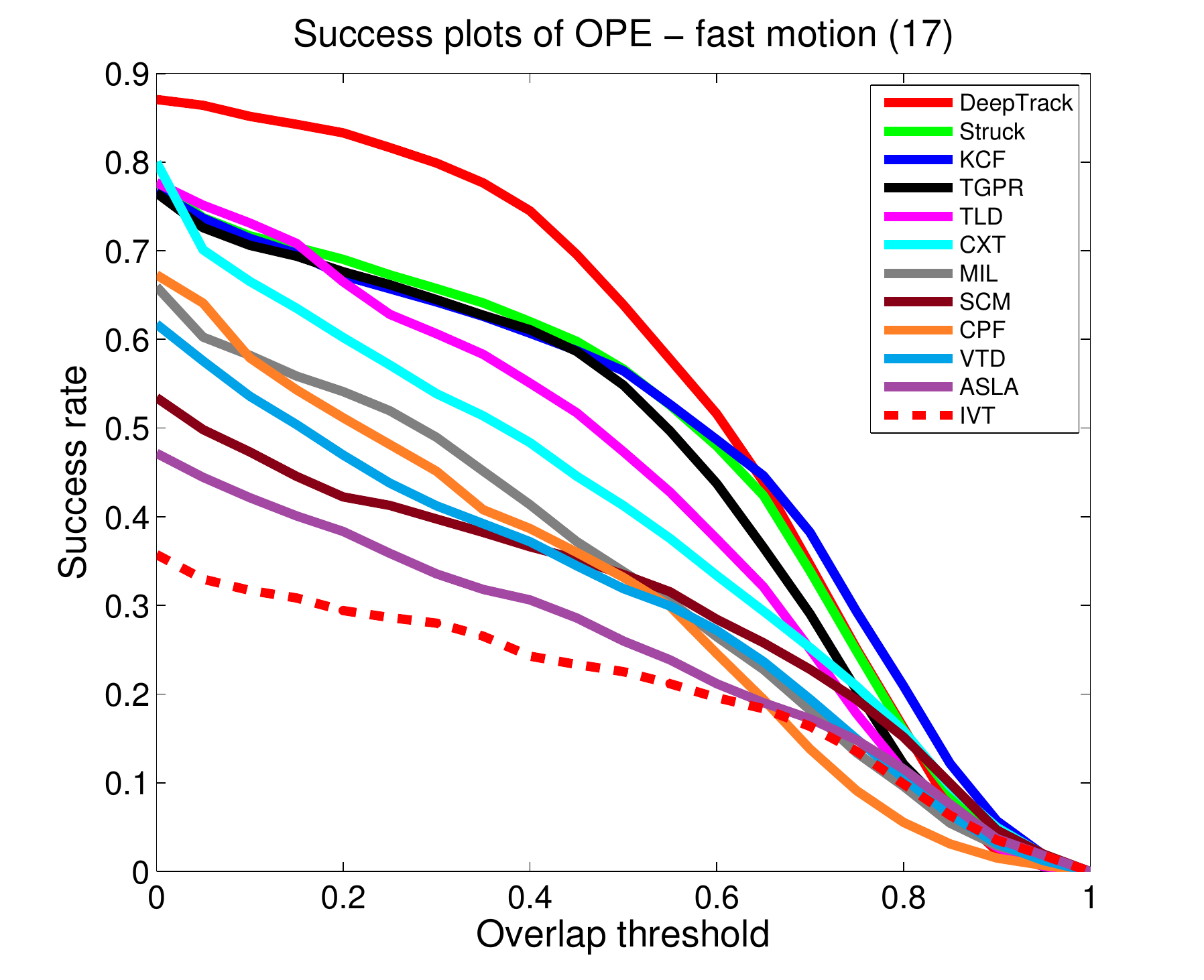}} 
  \subfigure{\includegraphics[width=0.24\textwidth]{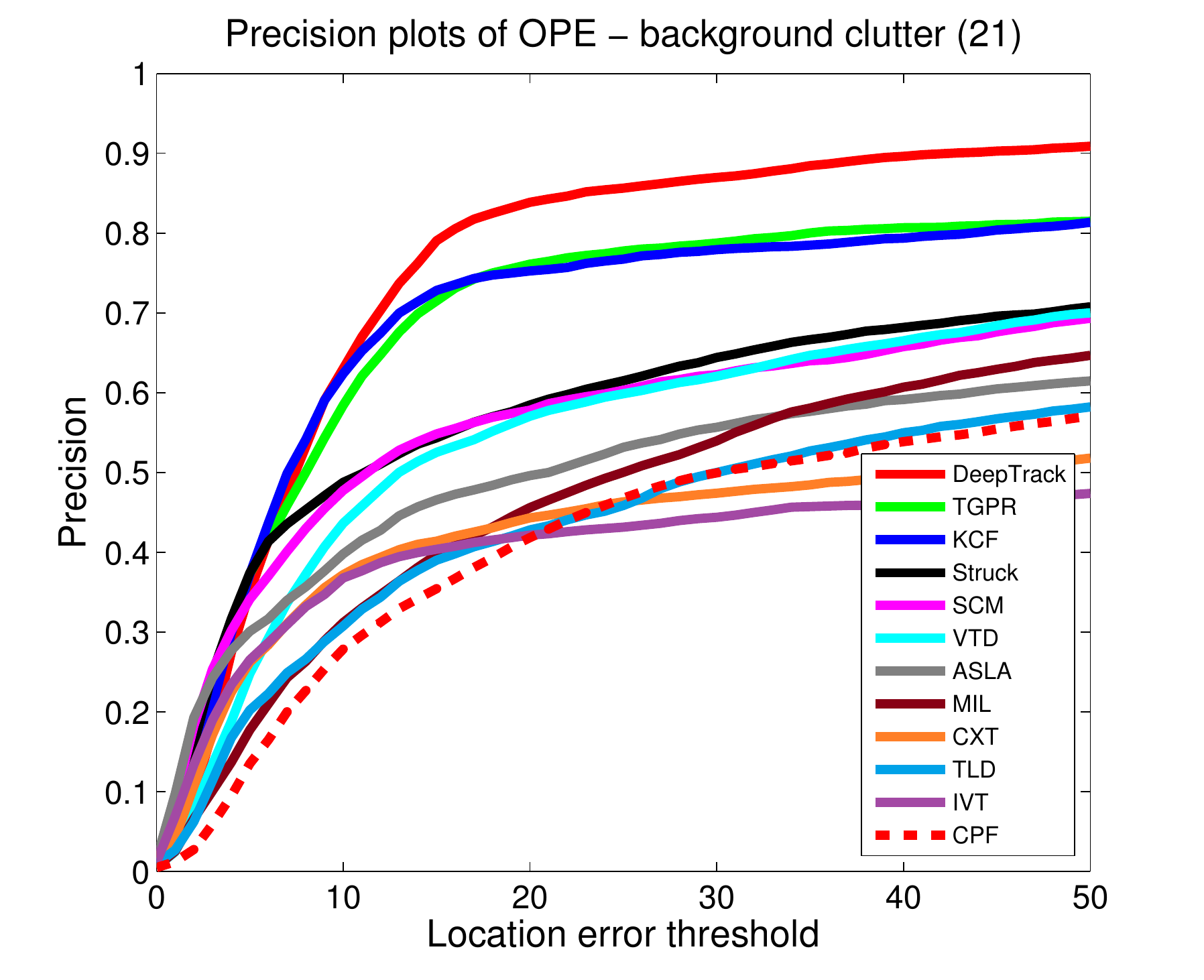}} 
  \subfigure{\includegraphics[width=0.24\textwidth]{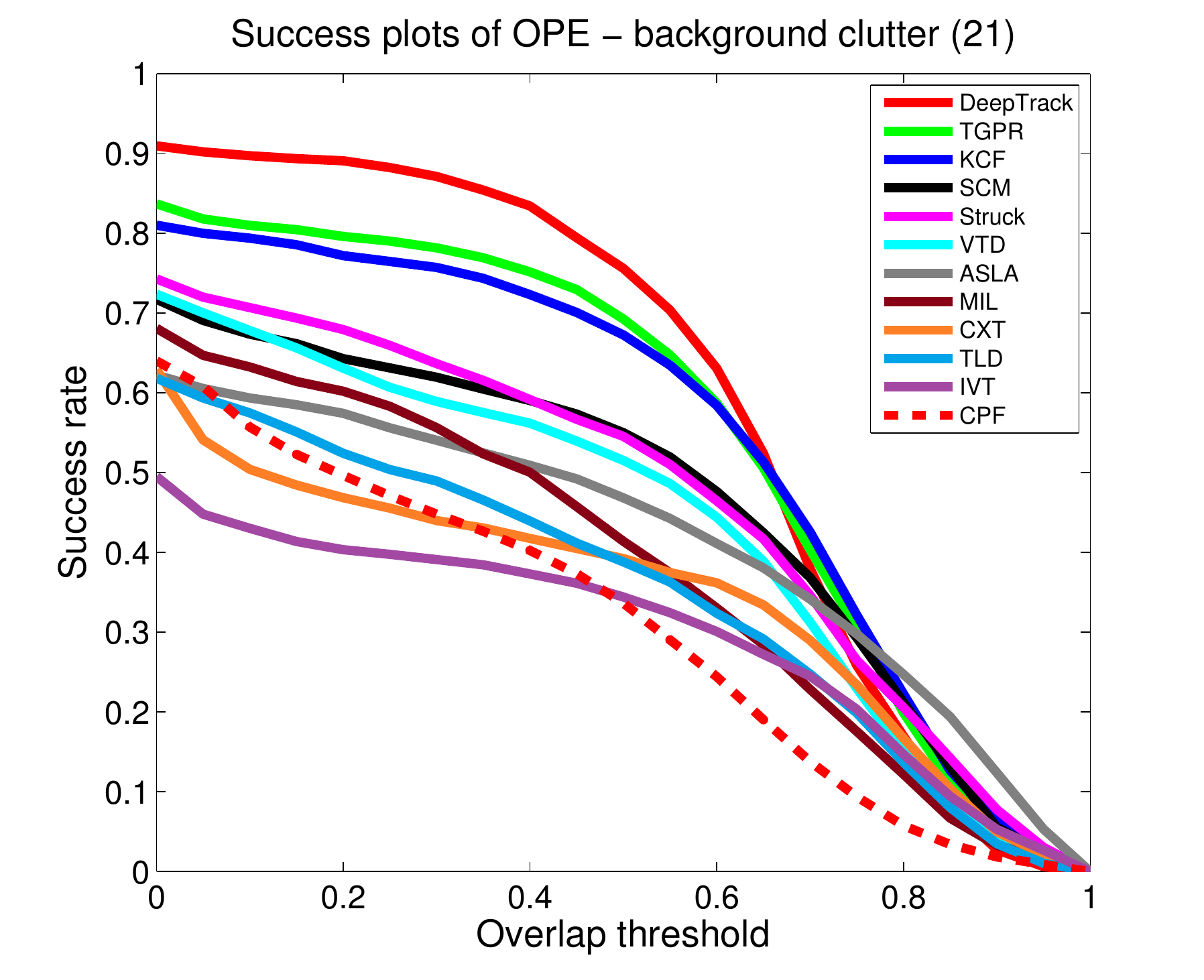}} 
  \subfigure{\includegraphics[width=0.24\textwidth]{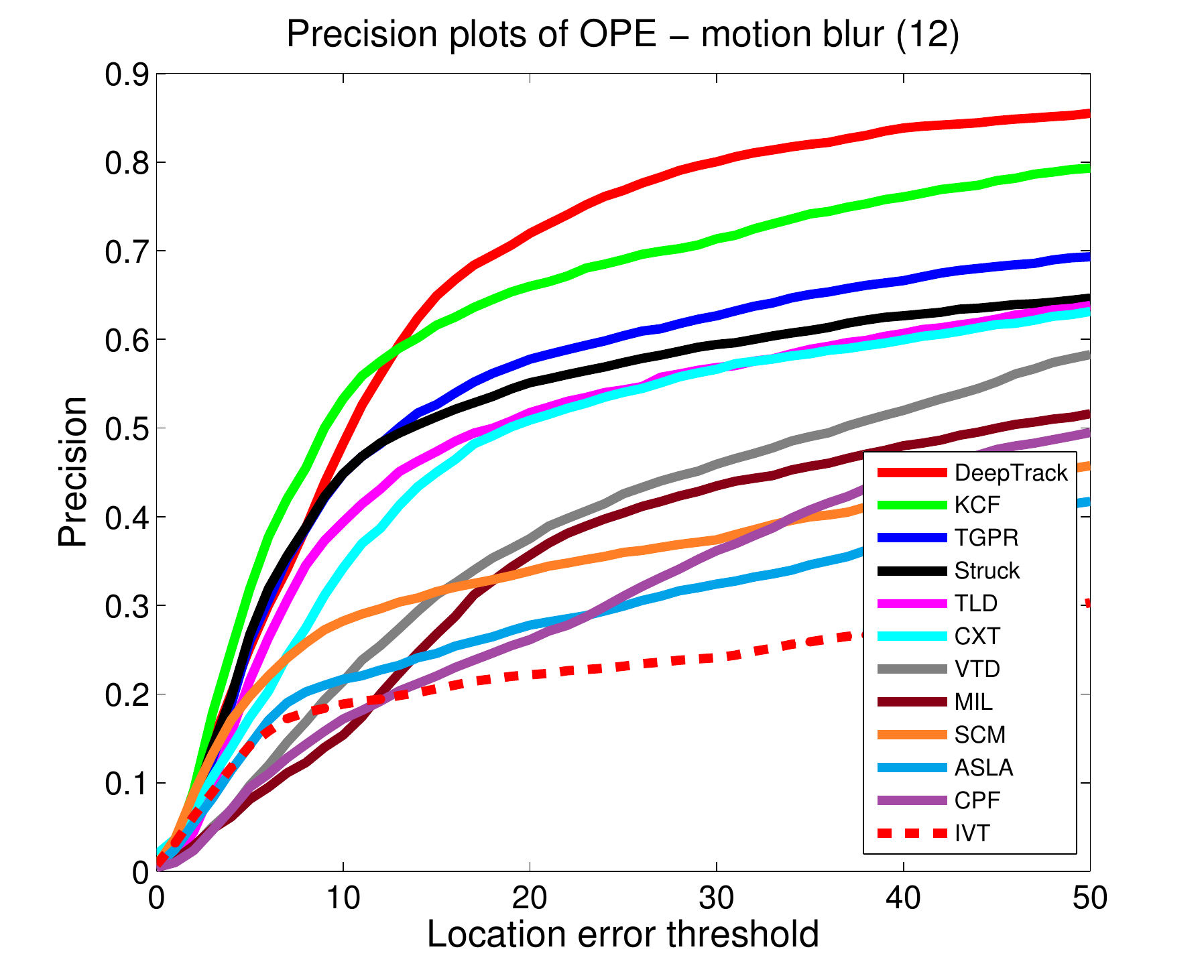}} 
  \subfigure{\includegraphics[width=0.24\textwidth]{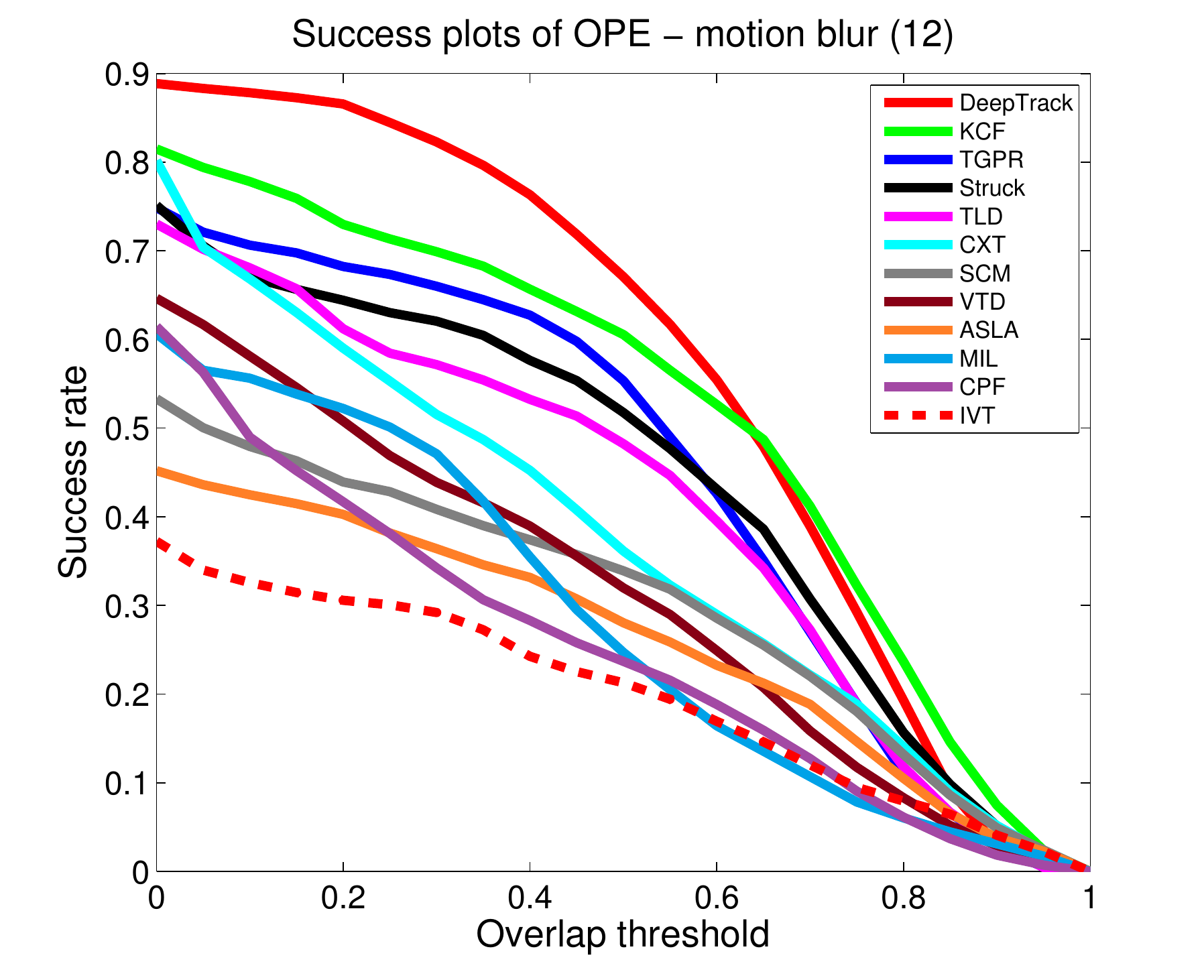}} 
  \subfigure{\includegraphics[width=0.24\textwidth]{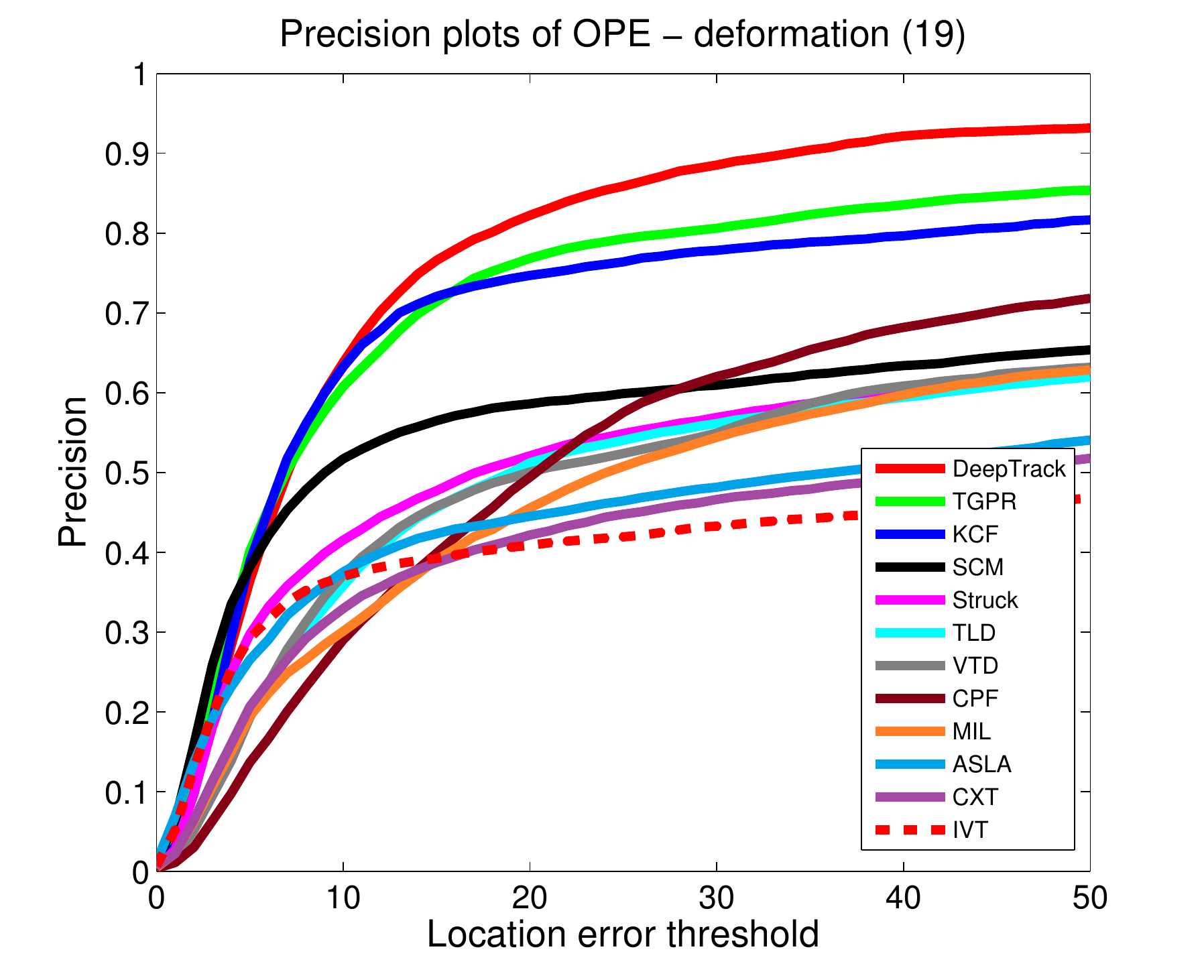}} 
  \subfigure{\includegraphics[width=0.24\textwidth]{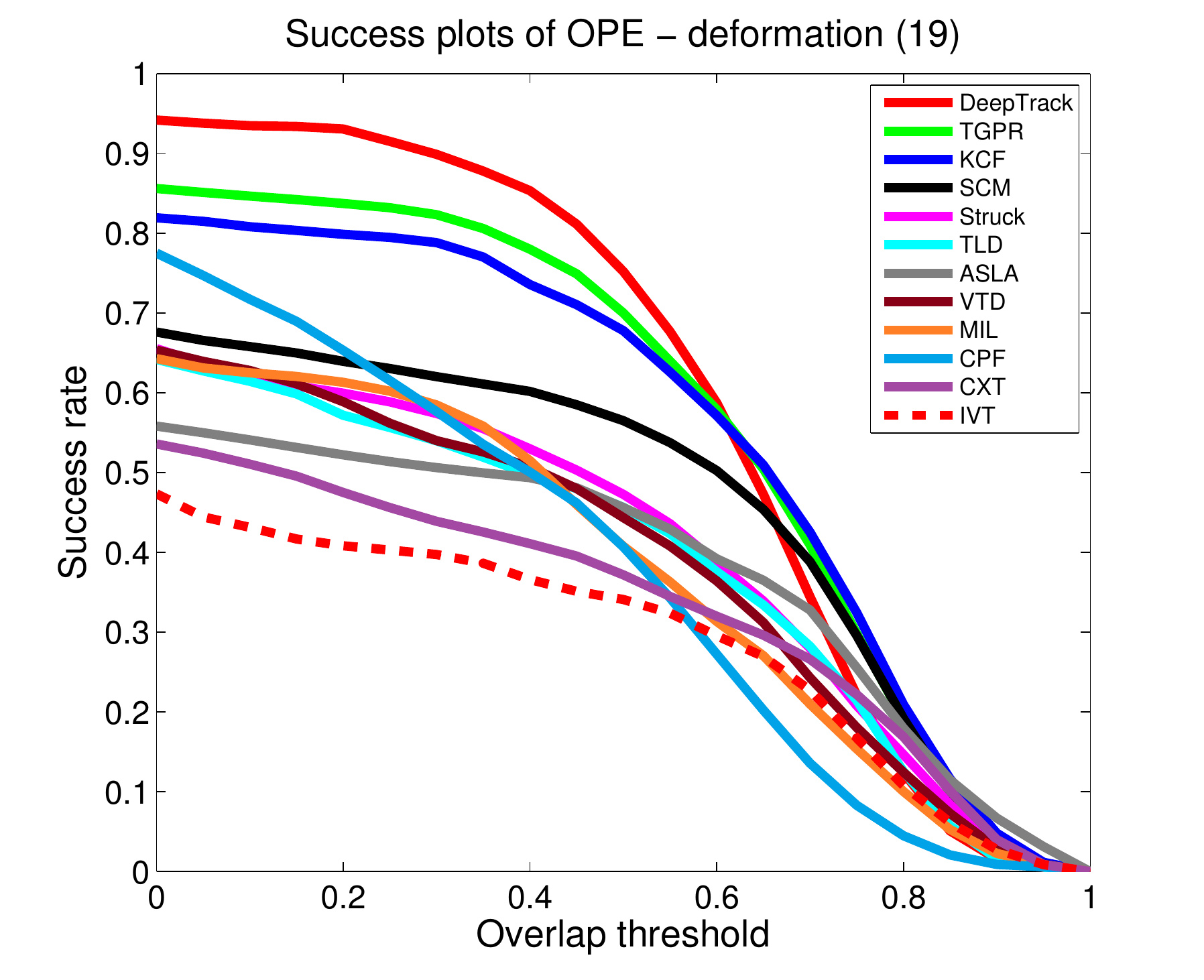}} 
  \subfigure{\includegraphics[width=0.24\textwidth]{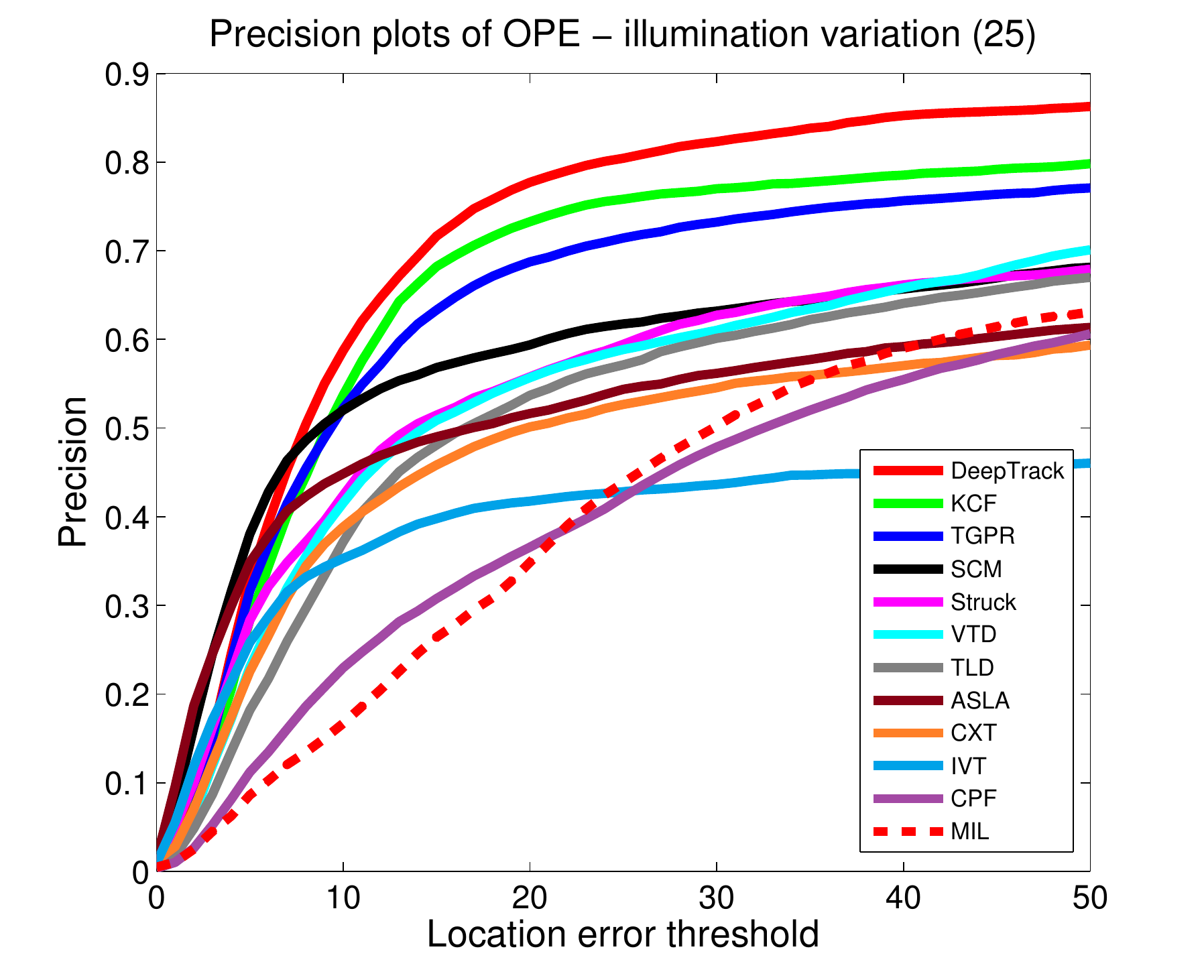}} 
  \subfigure{\includegraphics[width=0.24\textwidth]{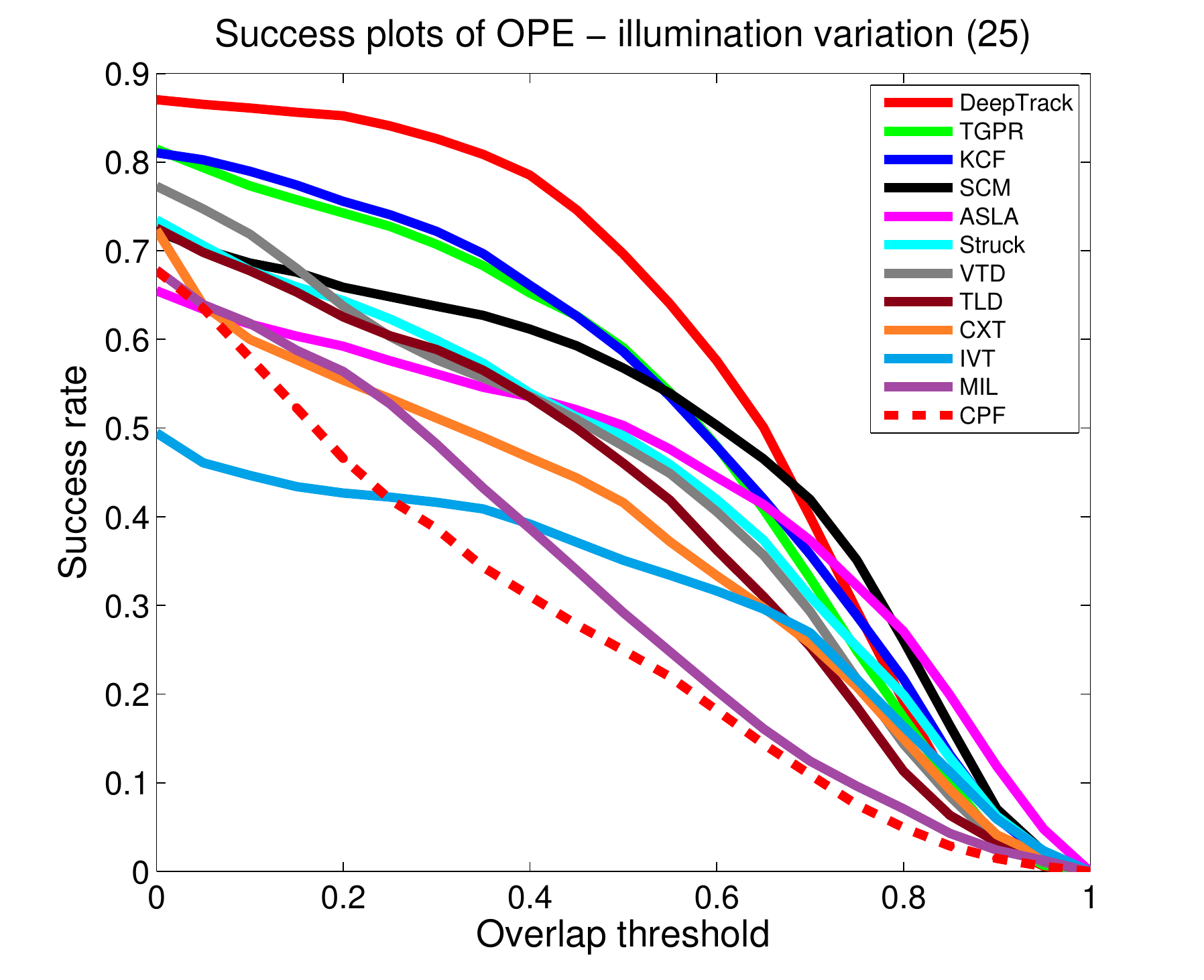}} 
  \subfigure{\includegraphics[width=0.24\textwidth]{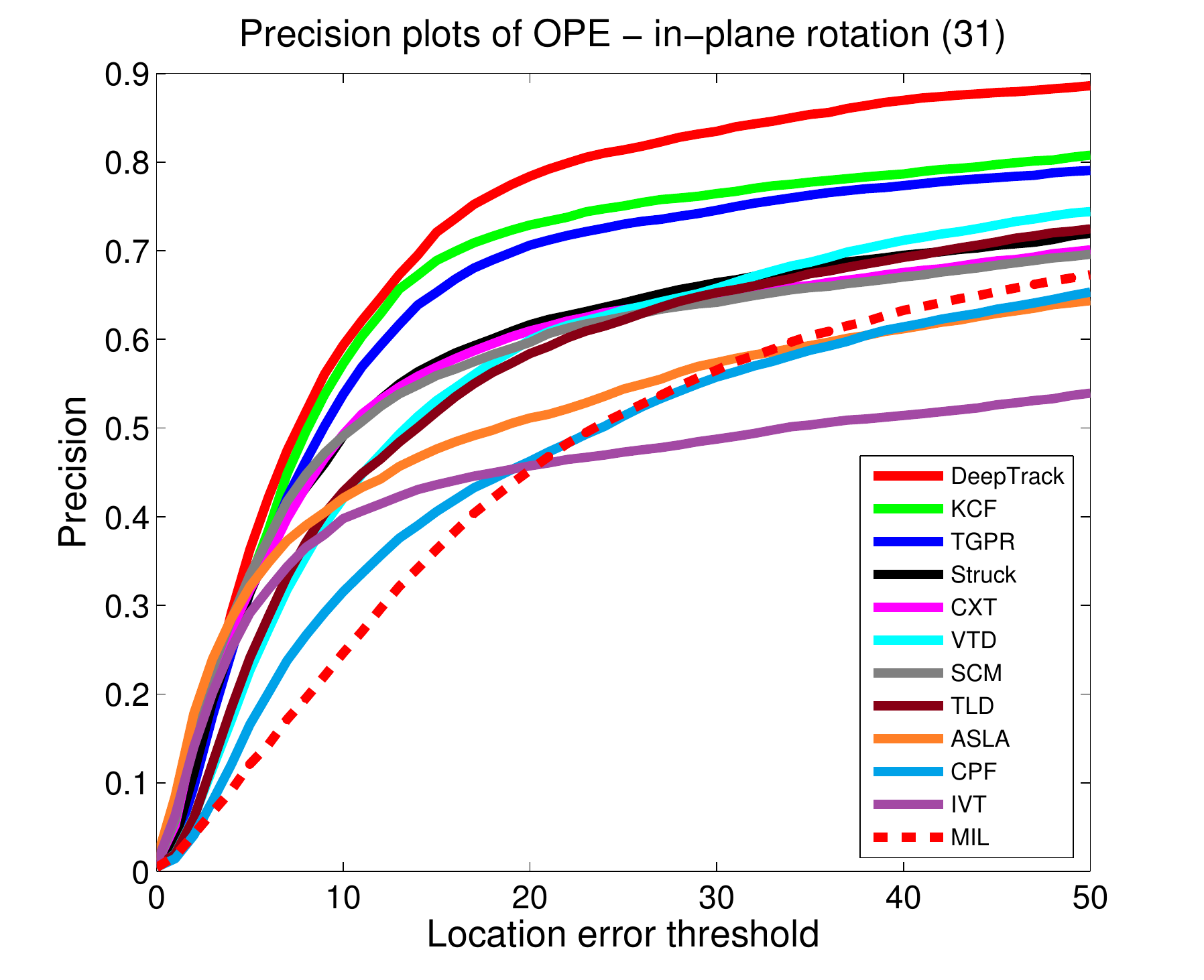}} 
  \subfigure{\includegraphics[width=0.24\textwidth]{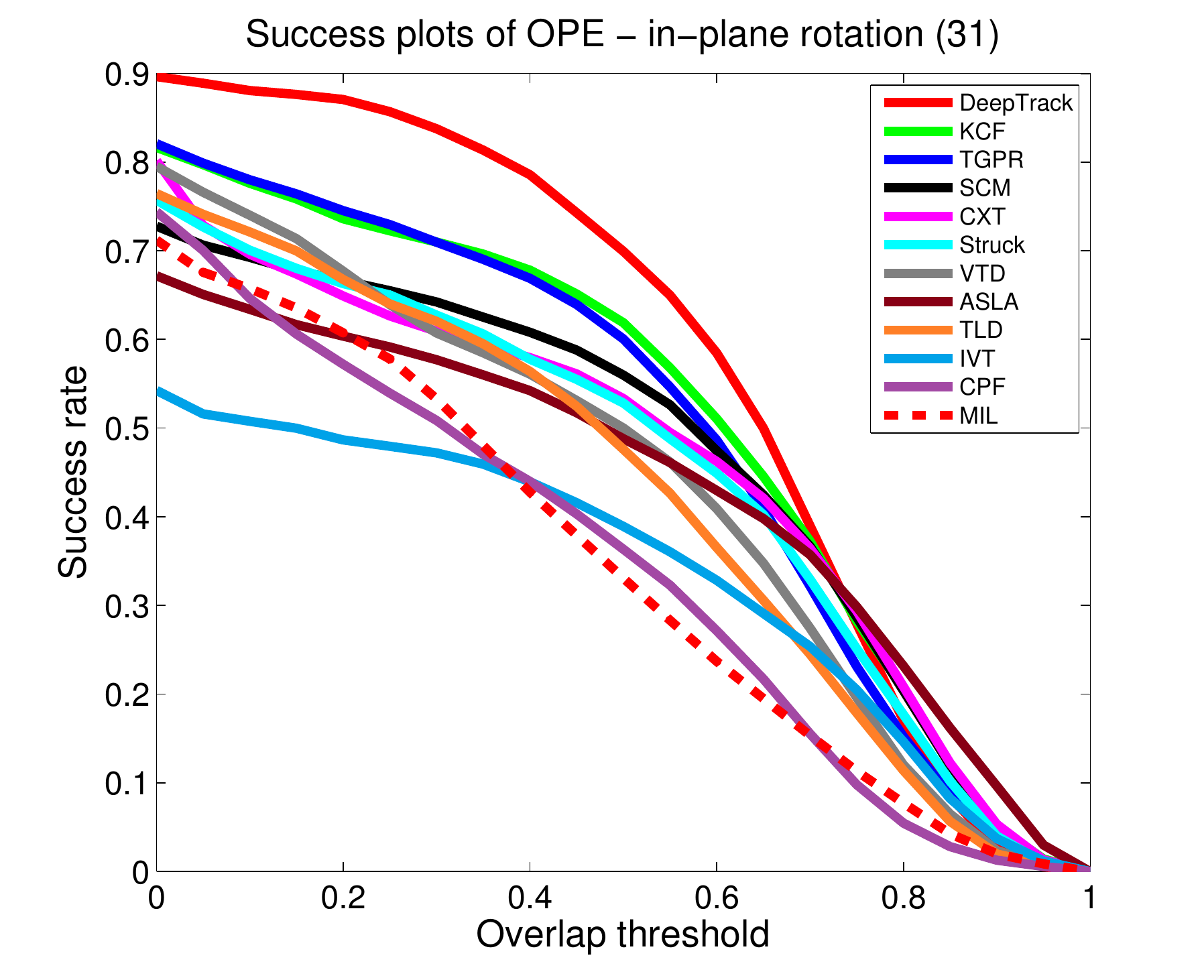}} 
  \subfigure{\includegraphics[width=0.24\textwidth]{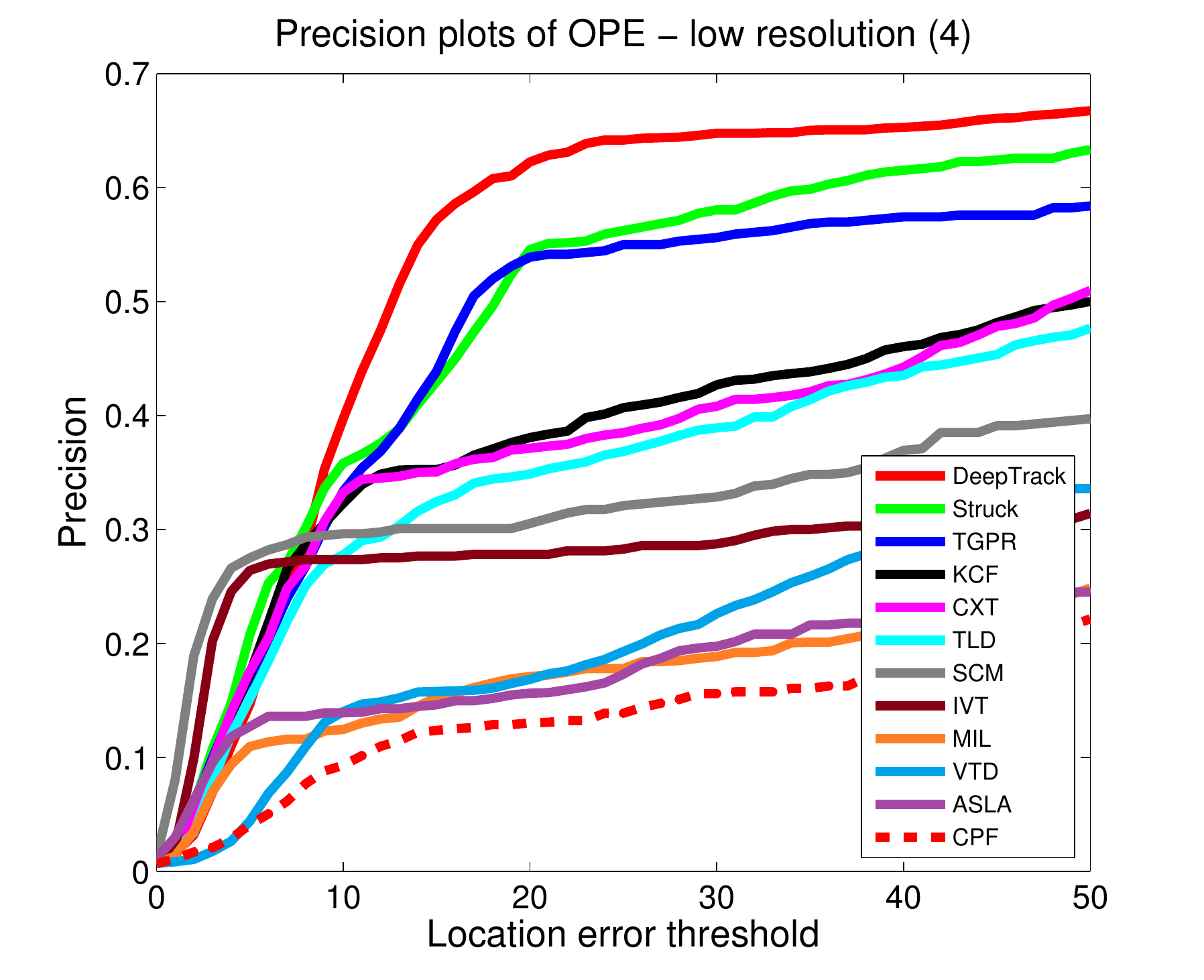}} 
  \subfigure{\includegraphics[width=0.24\textwidth]{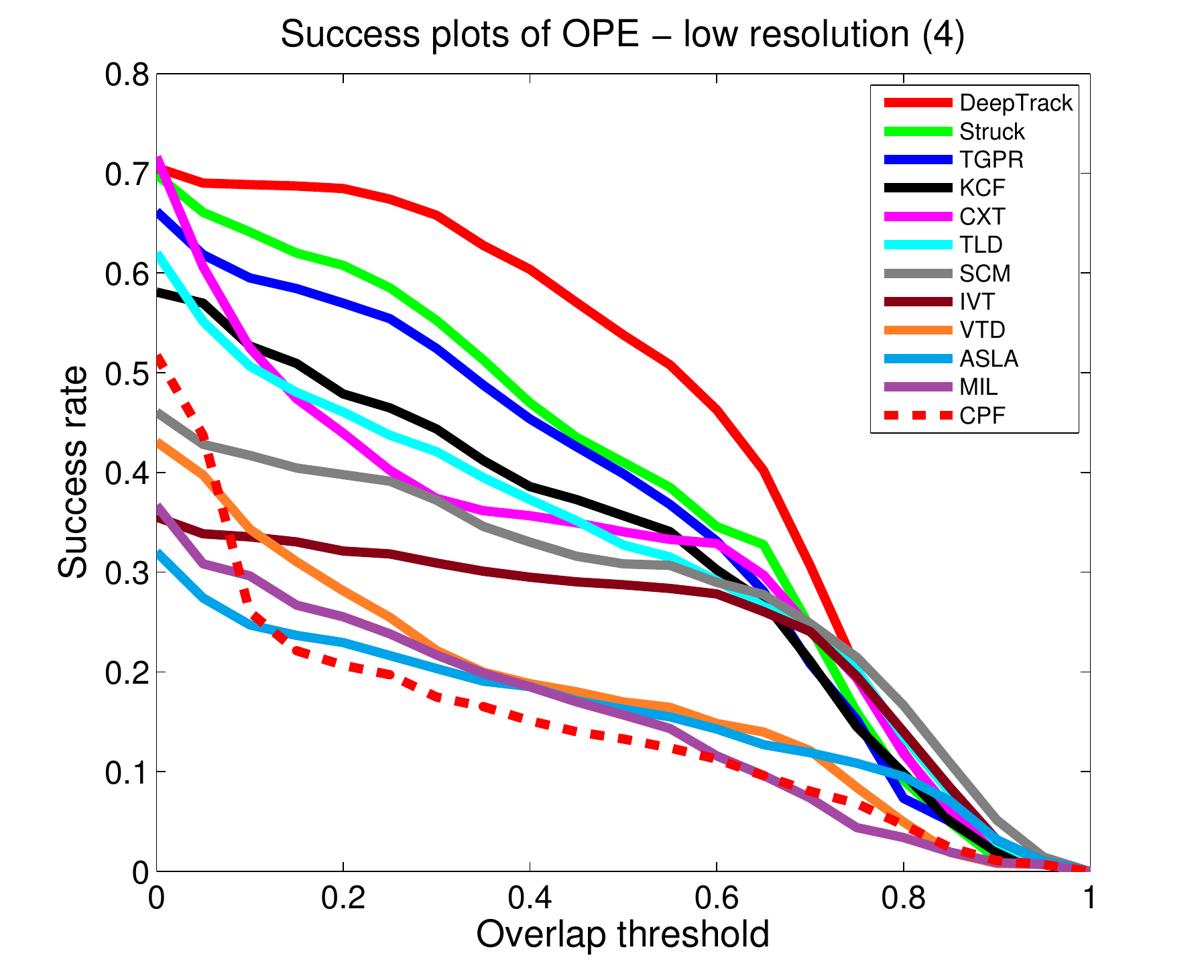}} 
  \subfigure{\includegraphics[width=0.24\textwidth]{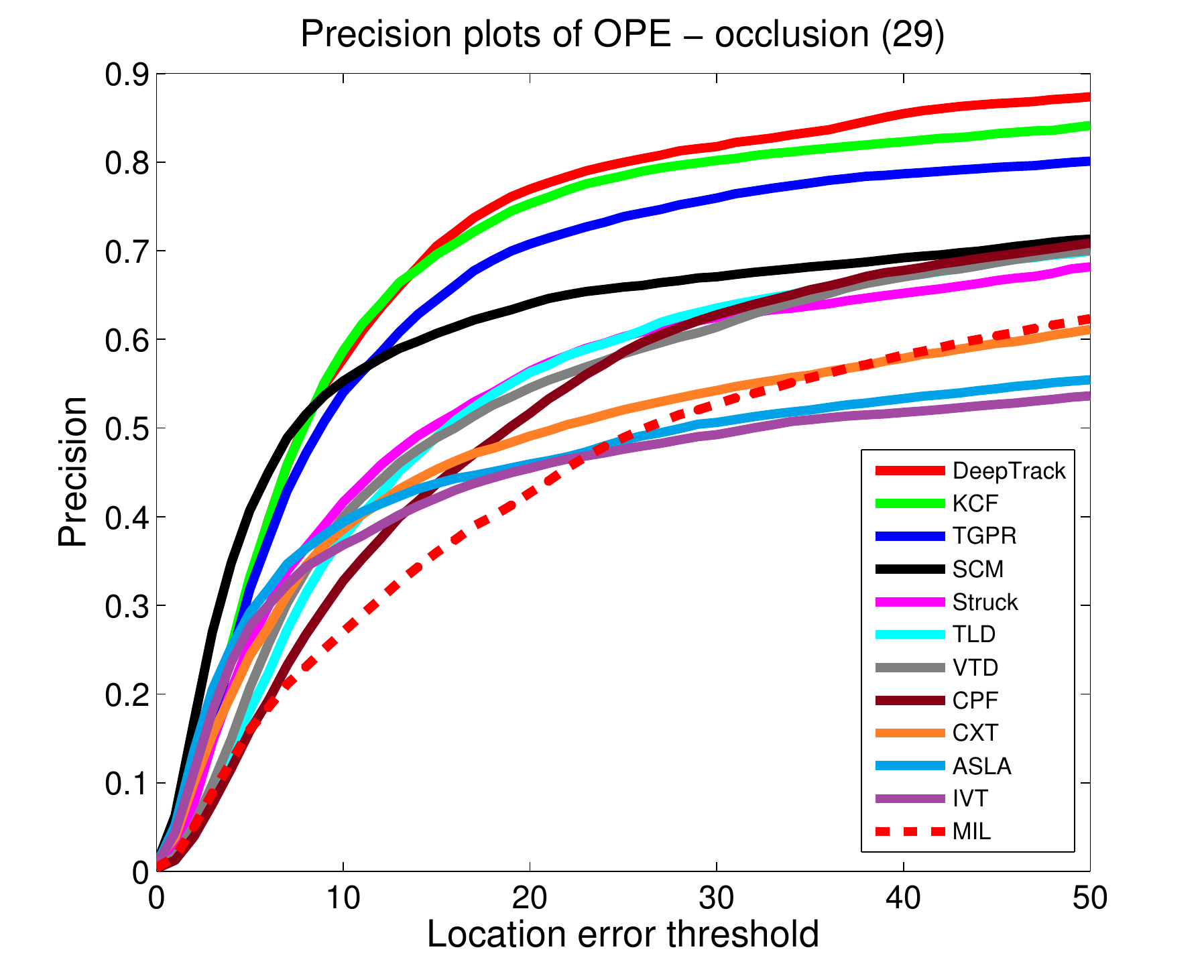}} 
  \subfigure{\includegraphics[width=0.24\textwidth]{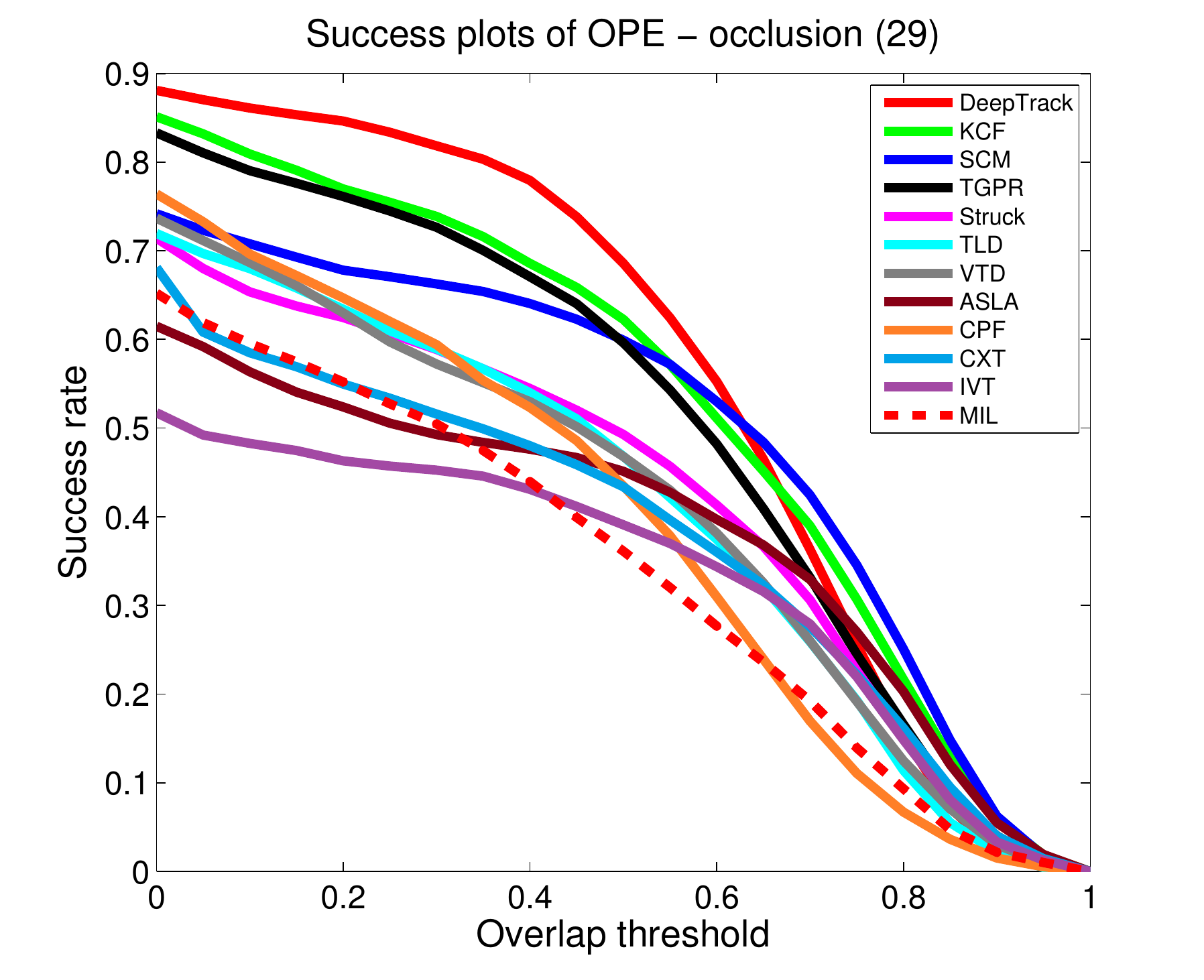}} 
  \subfigure{\includegraphics[width=0.24\textwidth]{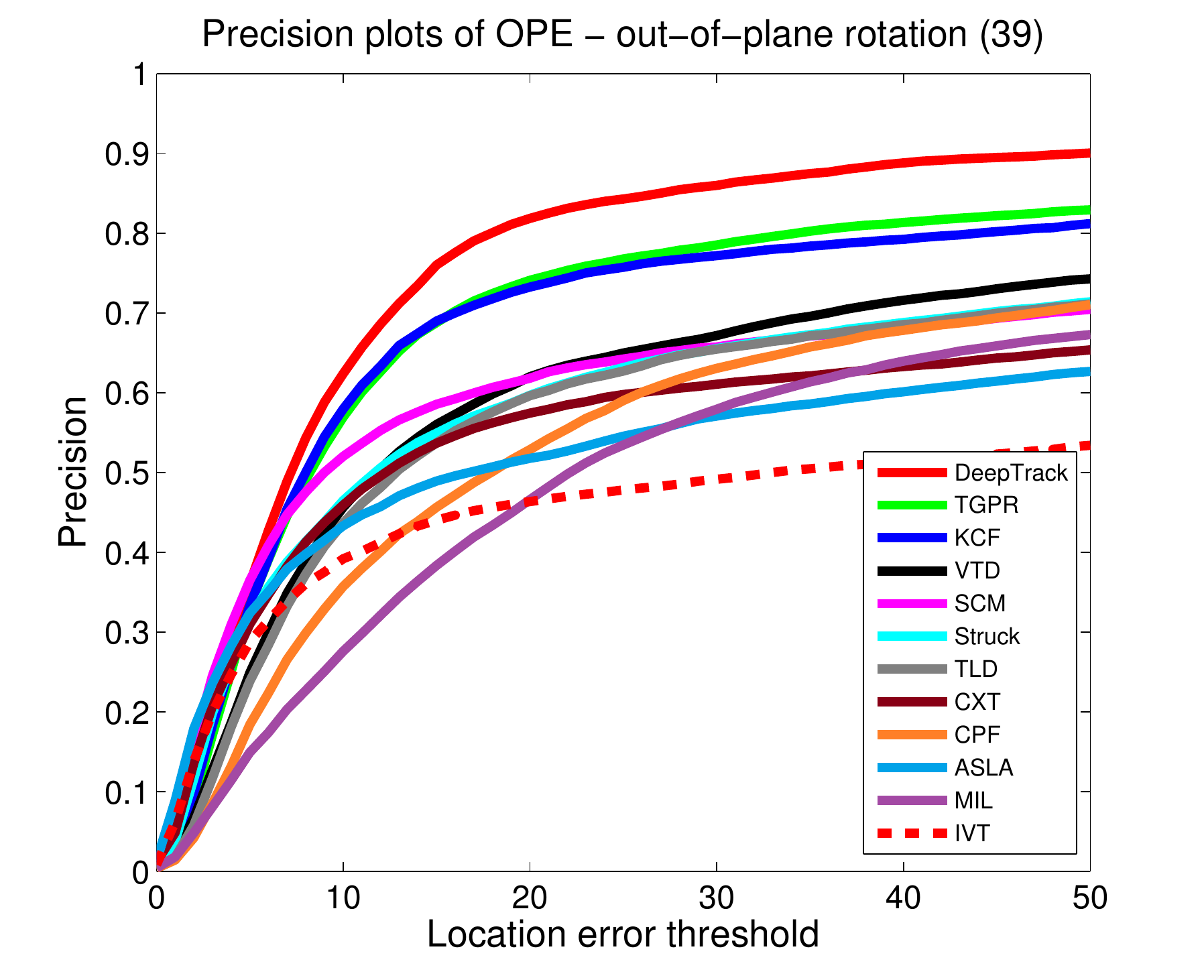}} 
  \subfigure{\includegraphics[width=0.24\textwidth]{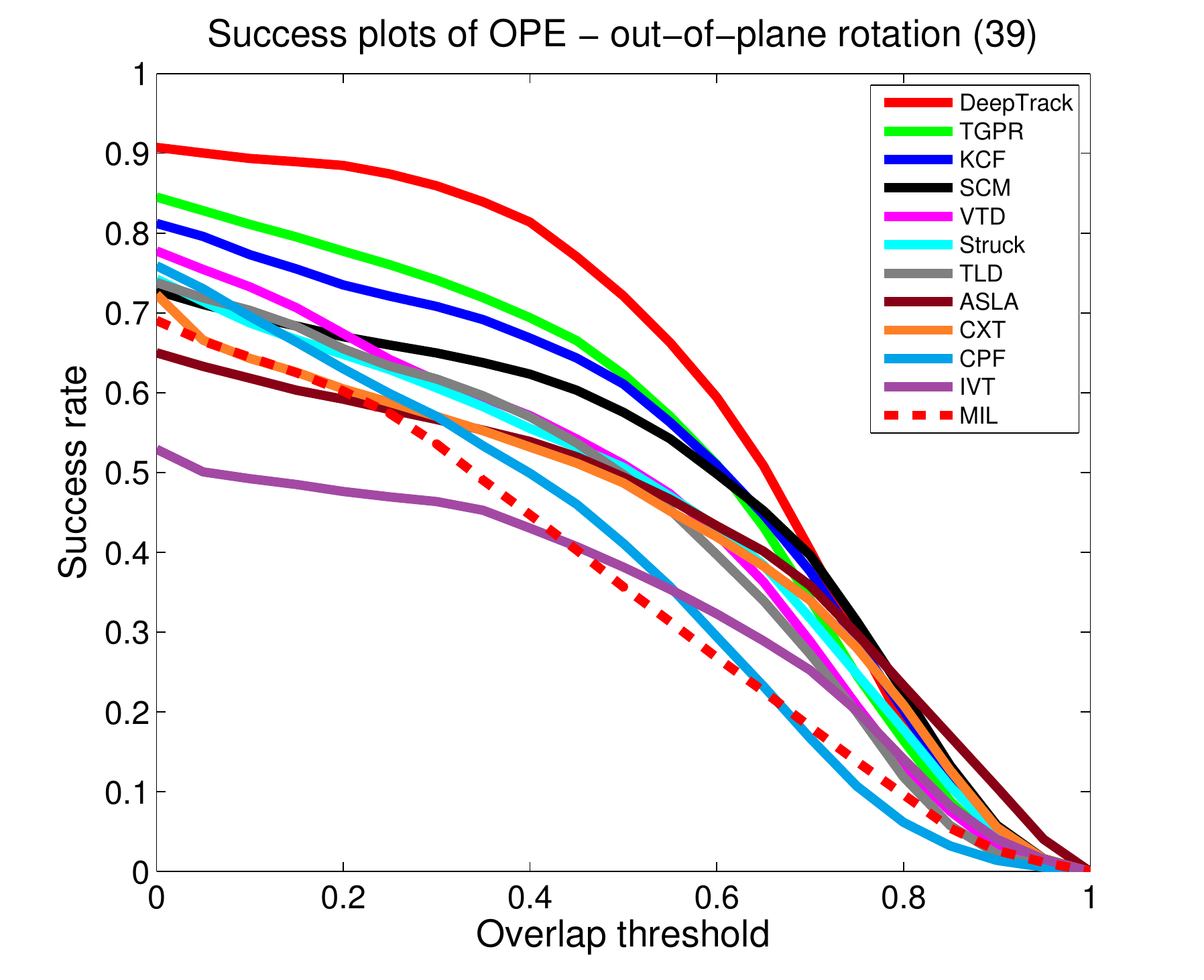}} 
  \subfigure{\includegraphics[width=0.24\textwidth]{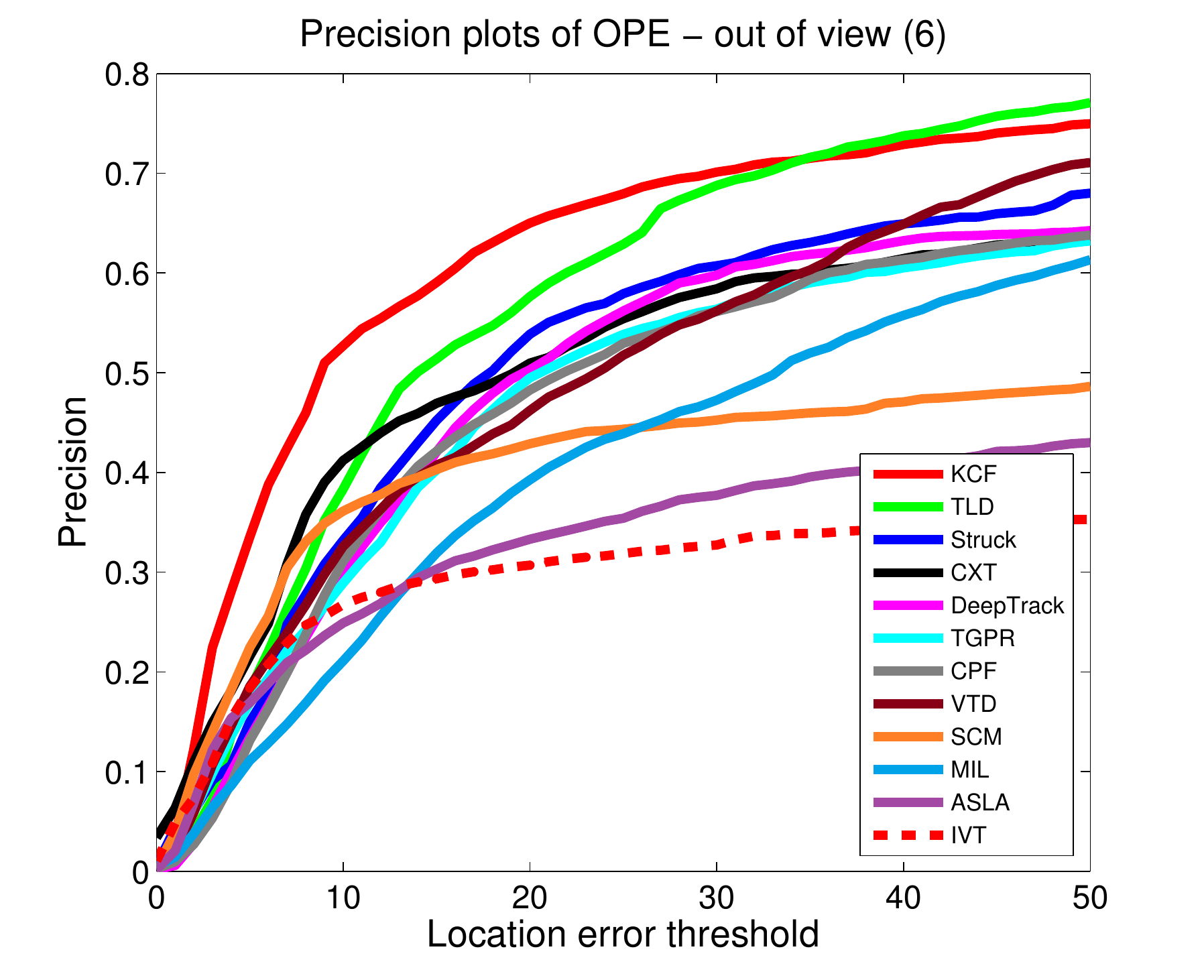}} 
  \subfigure{\includegraphics[width=0.24\textwidth]{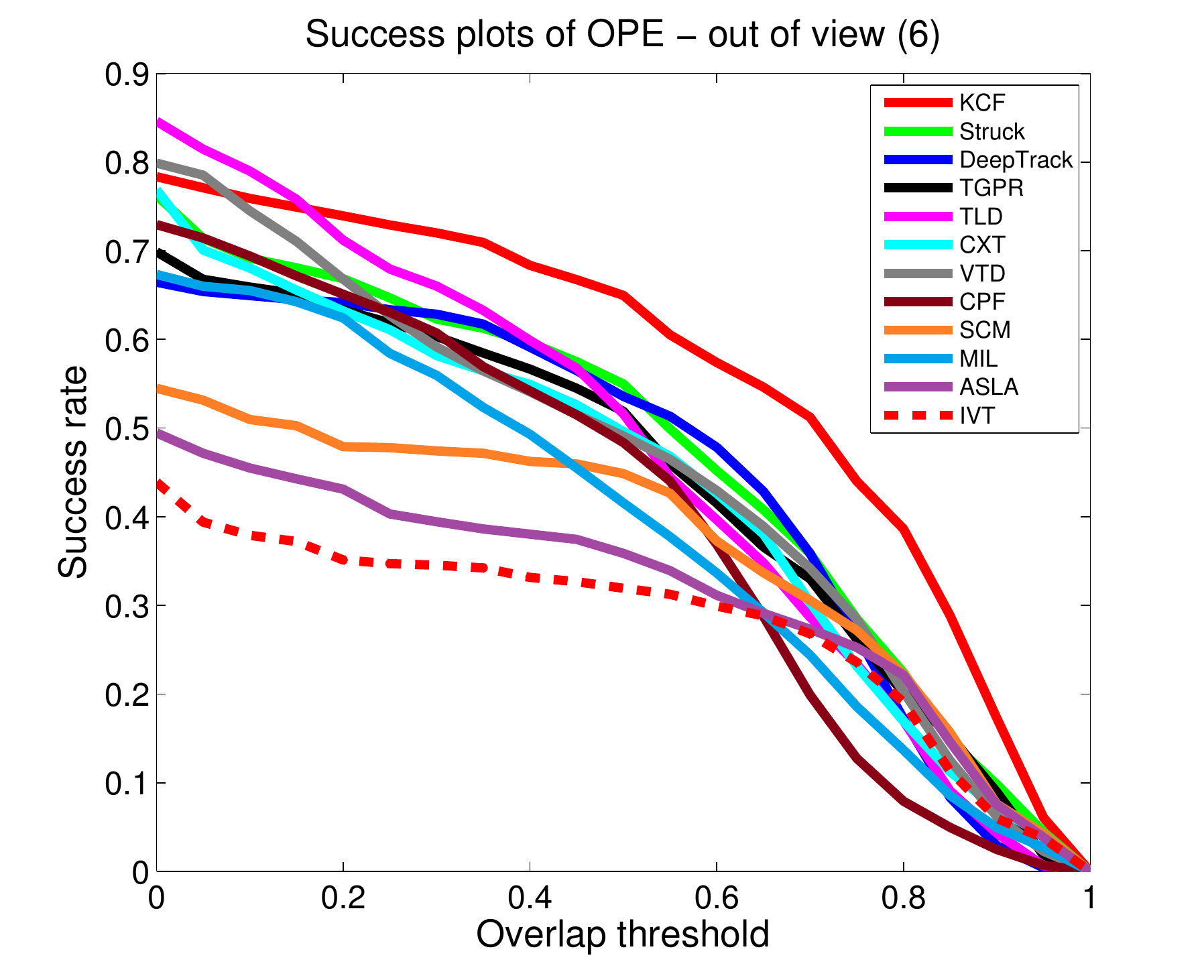}} 
  \subfigure{\includegraphics[width=0.24\textwidth]{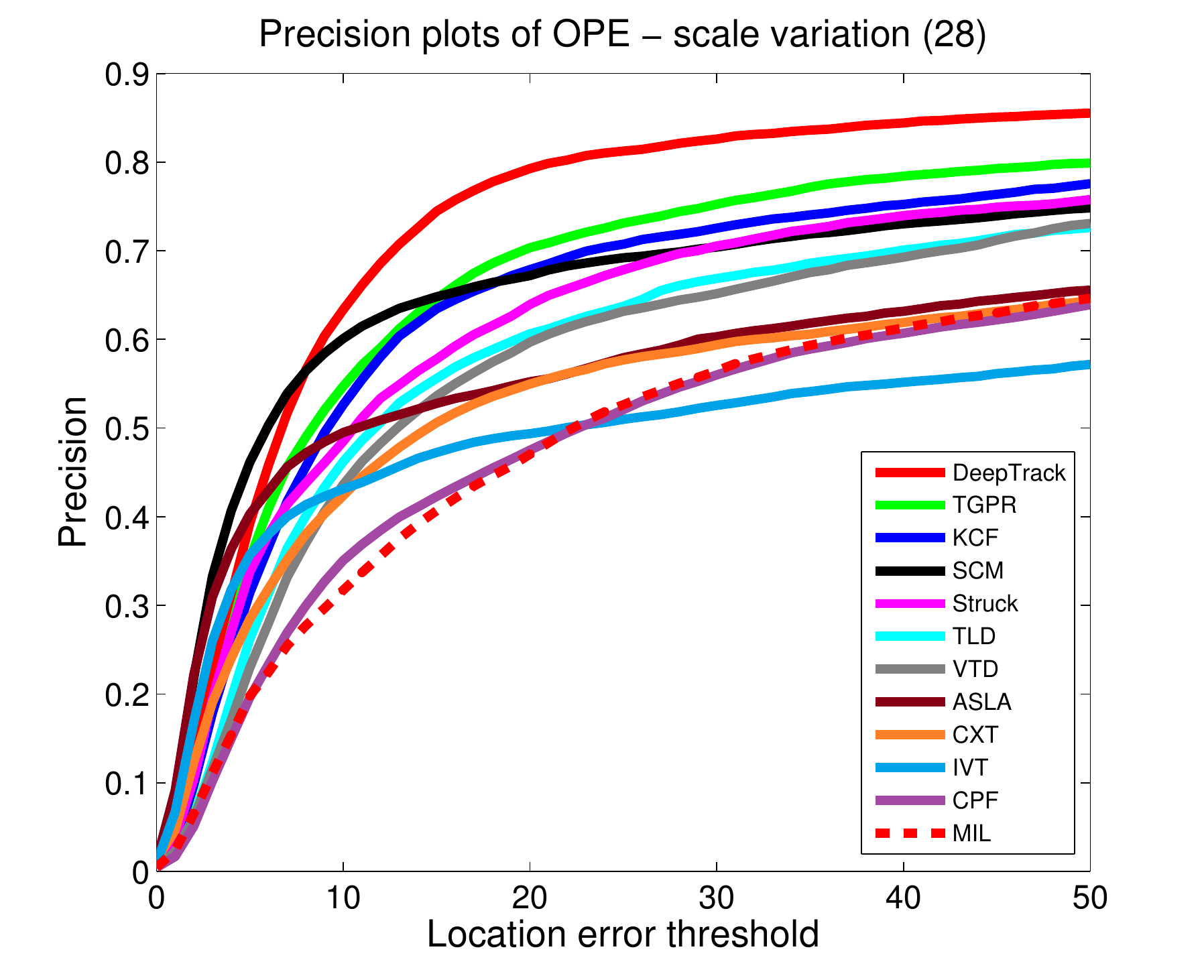}} 
  \subfigure{\includegraphics[width=0.24\textwidth]{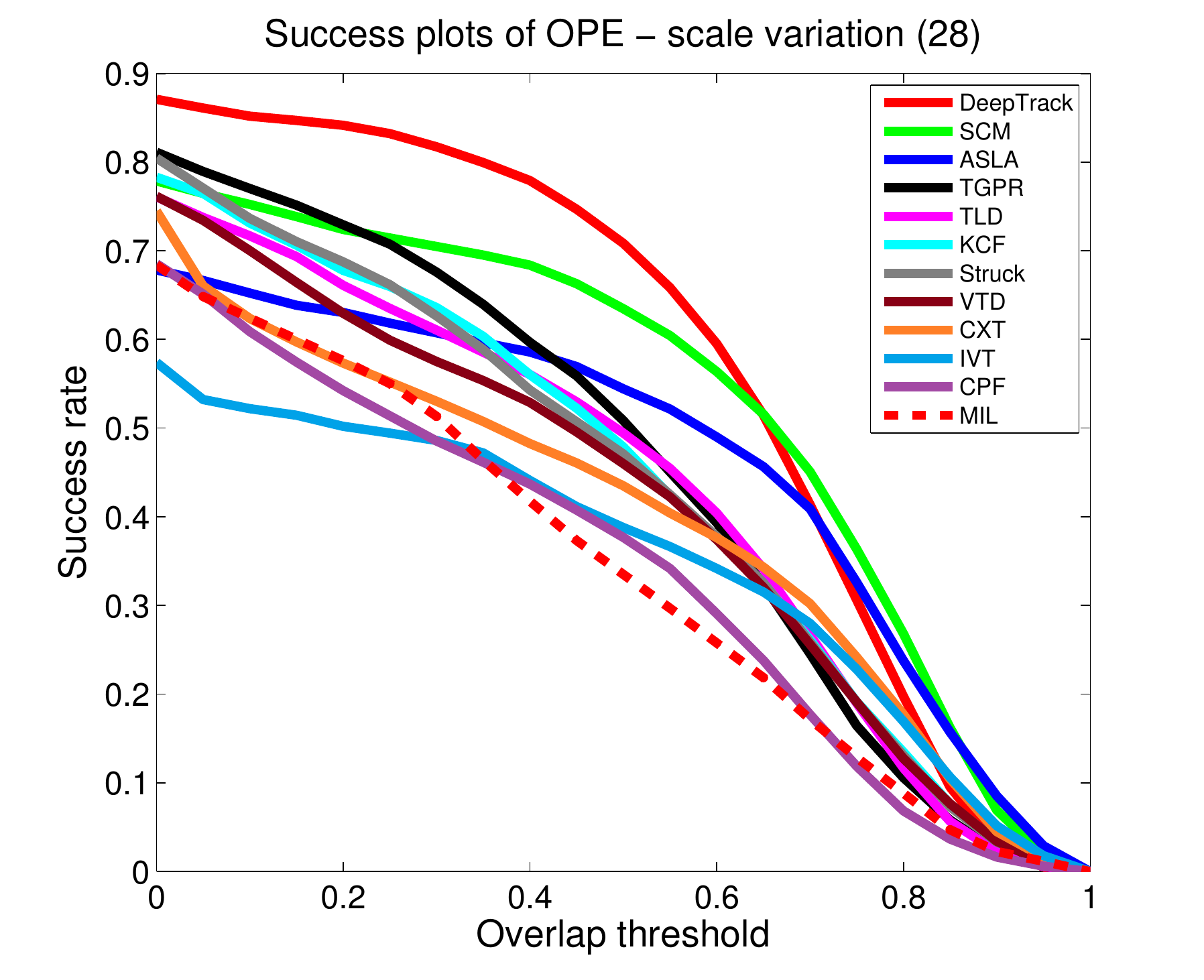}} 
  \caption
  {
    The Precision Plot (left) and the Success Plot (right) of the tracking results on the
    CVPR2013 benchmark, for $11$ kinds of tracking difficulties.
  }
\label{fig:cvpr13_attribuites}%

\end{figure*}

Fig.~\ref{fig:cvpr13_attribuites} shows the performance plots for $11$ kinds of
difficulties in visual tracking, {\it i.e.}, \emph{fast-motion, background-clutter,
motion-blur, deformation, illumination-variation, in-plane-rotation, low-resolution,
occlusion, out-of-plane-rotation, out-of-view and scale-variations}. We can see that the
proposed DeepTrack outperforms other competitors for all the difficulties except the
``out-of-view'' category.

\subsection{Comparison results on the VOT2013 benchmark}
\label{sec:comp2}

The VOT2013 Challenge Benchmark \cite{kristan2013visual} provides an evaluation kit and the dataset with 16 fully
annotated sequences for evaluating tracking algorithms in realistic scenes subject to
various common conditions. The tracking performance in the VOT2013 Challenge Benchmark is
primarily evaluated with two evaluation criteria: accuracy and robustness. The accuracy
measure is the average of the overlap ratios over the valid frames of each sequence while
the tracking robustness is the average number of failures over $15$ runs. A tracking
failure happens once the overlap ratio measure drops to zero and an re-initialization of
the tracker in the failure frame is conducted so it can continue. According to the
evaluation protocol, three types of experiments are conducted. In Experiment-$1$, the
tracker is run on each sequence in the dataset $15$ times by initializing it on the ground
truth bounding box. The setting of Experiment-$2$ is the same to Experiment-$1$, except
that the initial bounding box is randomly perturbed in the order of ten percent of the
object size. In Experiment-$3$, the colorful frames are converted into grayscale images.  

\begin{figure*}[ht!]
  \centering
  \subfigure{\includegraphics[width=0.085\textwidth]{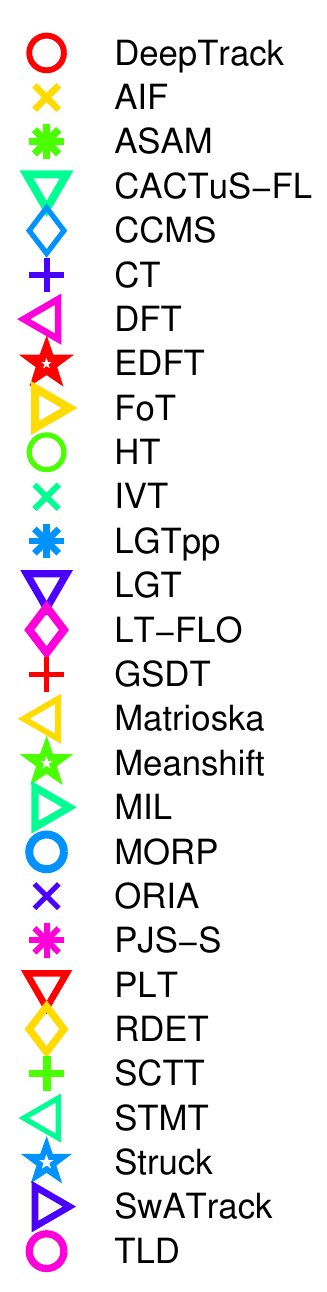}}
  \subfigure{\includegraphics[width=0.29\textwidth]{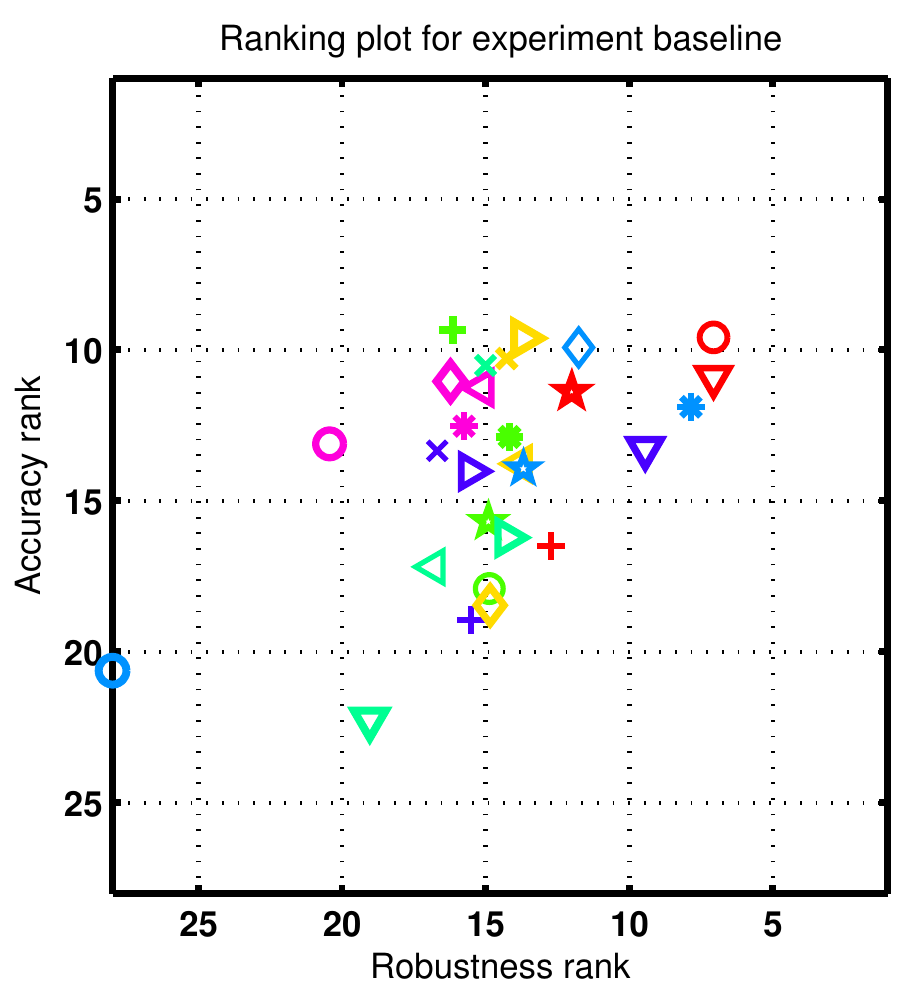}}
  \subfigure{\includegraphics[width=0.29\textwidth]{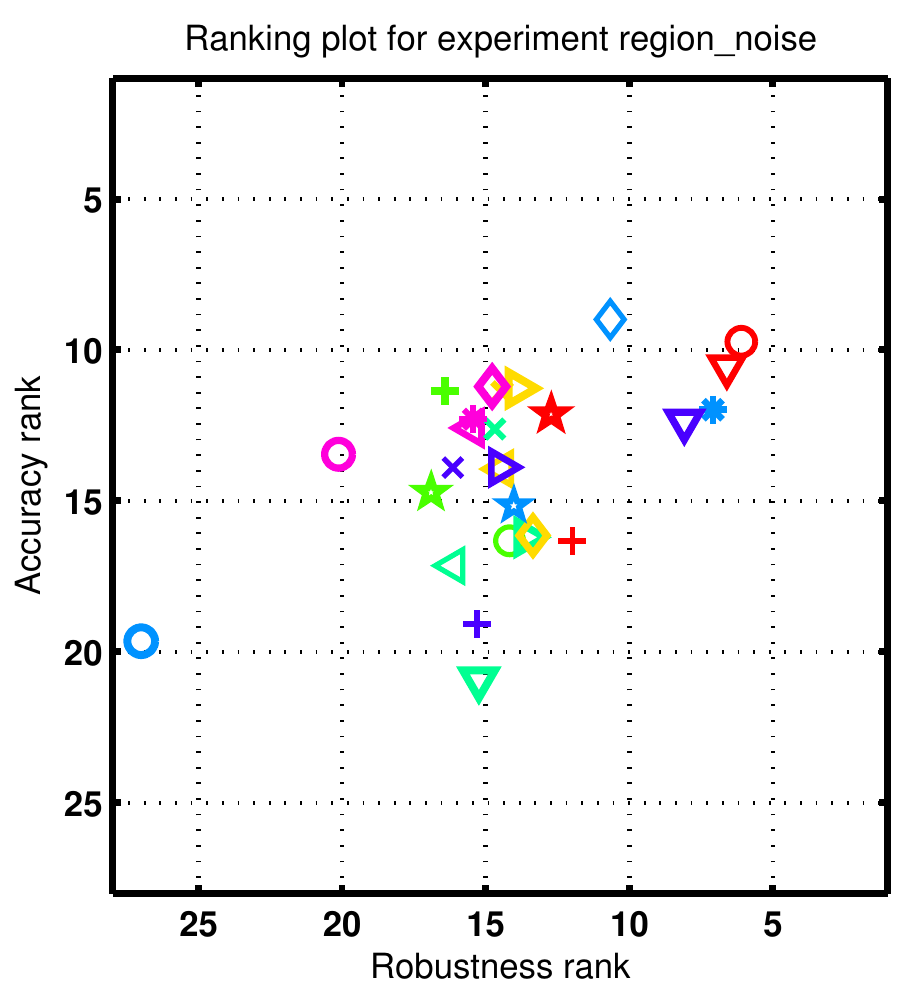}} 
  \subfigure{\includegraphics[width=0.29\textwidth]{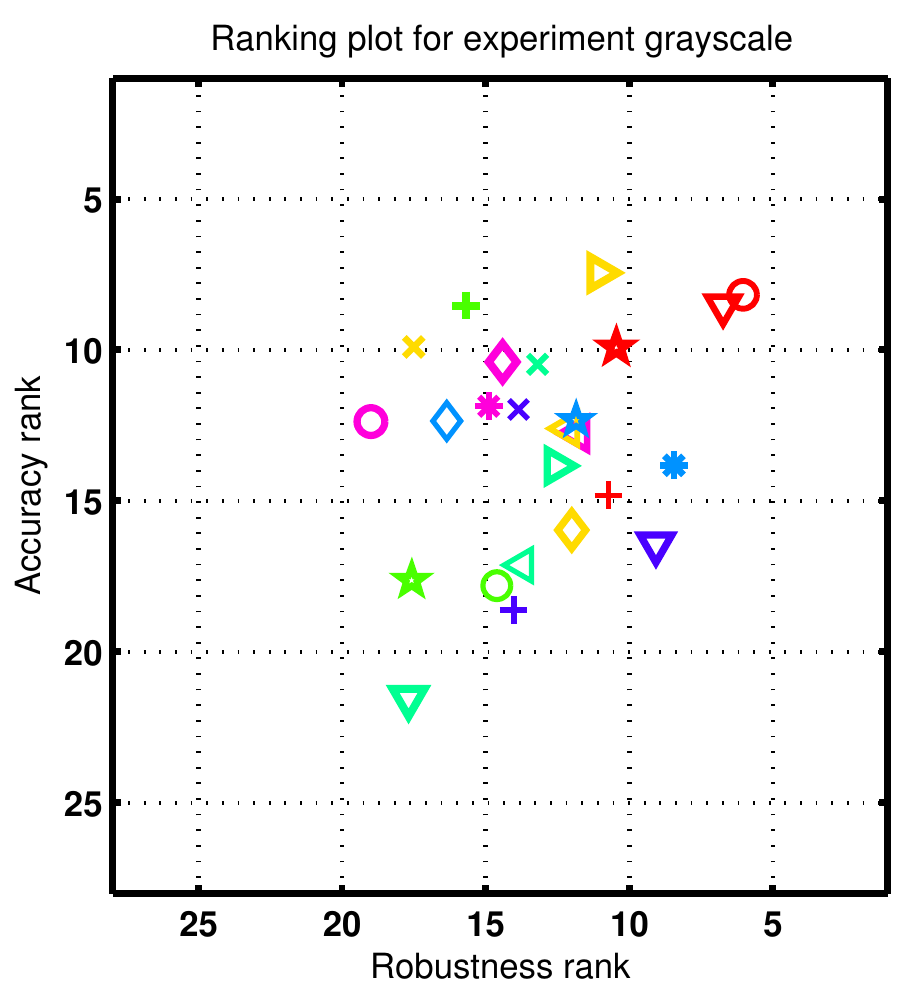}} \\
  \caption
  {
    The Precision Plot (left) and the Success Plot (right). The color of one curve is
    determined by the rank of the corresponding trackers, not their names. 
  }
\label{fig:vot_rank_plot}%
\end{figure*}

Firstly, we follow the evaluation protocol to test our method, compared with
other $27$ tracking algorithms provided in the benchmark website. The main comparison
results can be found in Table~\ref{tab:tracking_resutls_vot_all} and
Fig.~\ref{fig:vot_rank_plot}. We can see that, in average, the proposed method ranks the
first for both accuracy and robustness comparison. In specific, DeepTrack achieves the
best robustness scores for all the scenarios while ranks the second in accuracy for all
the experimental settings. In the Fig.~\ref{fig:vot_rank_plot}, one can observe that the
red circles (which stands for DeepTrack) always locate in the top-right corner of the
plot. This observation is consistent to the scores reported in
Table~\ref{tab:tracking_resutls_vot_all}. From the result we can see that our DeepTrack
achieves close while consistently better performances than the PLT method
\cite{kristan2013visual}. Other tracking methods that can achieve similar performances on
this benchmarks are FoT \cite{wendel2011robustifying}, EDFT \cite{felsberg2013enhanced}
and LGT++ \cite{xiao2013enhanced}.


\begin{table}[htb]
\centering
\resizebox{0.5\textwidth}{!}
{
  \begin{tabular}{ l | c | c | c | c | c | c | c | c }
  \hline\hline
  & \multicolumn{2}{|c|}{Experiment-1} & \multicolumn{2}{|c|}{Experiment-2}	&
  \multicolumn{2}{|c|}{Experiment-3}
  & \multicolumn{2}{|c}{Averaged} \\
  \cline{2-9}
  & Accu. & Rob.	& Accu. & Rob. & Accu. & Rob. & Accu. & Rob. \\
  \cline{1-9}
  CNN	& $\color{blue}9.60$ & $\color{red}\bf 7.06$ & $\color{blue}10.14$ & $\color{red}\bf 6.09$ &
        $\color{blue}8.17$ & $\color{red}\bf 6.04$ & $\color{red}\bf 9.30$ & $\color{red}\bf 6.40$ \\
  AIF & $10.29$ & $14.27$ & $11.39$ & $14.43$ & $9.90$ & $17.50$ & $10.52$ & $15.40$ \\
  ASAM & $12.86$ & $14.17$ & NaN & NaN & NaN & NaN & NaN & NaN \\
  CACTuS-FL	& $22.27$	& $19.03$ & $20.92$ & $15.24$ & $21.54$ & $17.69$ & $21.58$ & $17.32$ \\
  CCMS & $9.87$ & $11.76$ & $\color{red}\bf 8.97$ & $10.66$ & $12.36$ & $16.35$ & $10.40$ & $12.92$ \\
  CT & $18.92$ & $15.51$ & $19.10$ & $15.30$ & $18.62$ & $14.03$ & $18.88$ & $14.95$ \\
  DFT & $11.23$ & $15.12$ & $12.64$ & $15.47$ & $12.78$ & $11.79$ & $12.21$ & $14.13$ \\
  EDFT & $11.45$ & $12.00$ & $12.18$ & $12.72$ & $9.91$ & $10.44$ & $11.18$ & $11.72$ \\
   FoT & $9.63$ & $13.69$ & $11.23$ & $13.87$ & $\color{red}\bf 7.44$ & $11.04$ & $\color{blue}9.44$ & $12.87$ \\
  HT & $17.90$ & $14.88$ & $16.29$ & $14.17$ & $17.81$ & $14.61$ & $17.33$ & $14.55$ \\
  IVT & $10.51$ & $15.00$ & $12.66$ & $14.68$ & $10.48$ & $13.19$ & $11.21$ & $14.29$ \\
LGTpp & $11.89$ & $\color{blue}7.84$ & $11.98$ & $7.08$ & $13.82$ & $8.44$ & $12.56$ & $7.79$ \\
  LGT & $13.27$ & $9.44$ & $12.37$ & $8.08$ & $16.43$ & $9.07$ & $14.02$ & $8.86$ \\
  LT-FLO & $11.09$ & $16.22$ & $11.25$ & $14.77$ & $10.40$ & $14.41$ & $10.91$ & $15.14$ \\
  GSDT & $16.51$ & $12.73$ & $16.36$ & $11.98$ & $14.82$ & $10.73$ & $15.90$ & $11.81$ \\
  Matrioska & $13.77$ & $13.78$ & $13.94$ & $14.43$ & $12.60$ & $12.13$ & $13.44$ & $13.45$ \\
  Meanshift & $15.69$ & $14.91$ & $14.82$ & $16.90$ & $17.64$ & $17.57$ & $16.05$ & $16.46$ \\
  MIL & $16.38$ & $14.25$ & $16.28$ & $13.58$ & $13.84$ & $12.54$ & $15.50$ & $13.46$ \\
  MORP & $20.64$ & $28.00$ & $19.65$ & $27.00$ & NaN & NaN & NaN & NaN \\
  ORIA & $13.13$ & $16.69$ & $13.86$ & $16.15$ & $11.97$ & $13.85$ & $12.99$ & $15.56$ \\
  PJS-S & $12.50$ & $15.75$ & $12.31$ & $15.43$ & $11.87$ & $14.89$ & $12.22$ & $15.36$ \\
    PLT & $10.88$ & $\color{red}\bf 7.06$ & $10.58$ & $\color{blue}6.60$ & $8.54$ &
    $\color{blue}6.73$ & $10.00$ & $\color{blue}6.79$ \\
  RDET & $18.42$ & $14.84$ & $16.14$ & $13.35$ & $15.97$ & $12.00$ & $16.84$ & $13.40$ \\
  SCTT & $\color{red}\bf 9.36$ & $16.16$ & $11.37$ & $16.43$ & $8.53$ & $15.68$ & $9.75$ & $16.09$ \\
  STMT & $17.16$ & $16.81$ & $17.17$ & $16.12$ & $17.12$ & $13.73$ & $17.15$ & $15.55$ \\
  Struck & $13.92$ & $13.69$ & $15.21$ & $14.02$ & $12.33$ & $11.85$ & $13.82$ & $13.19$ \\
  SwATrack & $13.98$ & $15.53$ & $13.93$ & $14.48$ & NaN & NaN & NaN & NaN \\
  TLD & $13.12$ & $20.44$ & $13.37$ & $20.12$ & $12.37$ & $19.00$ & $12.95$ & $19.85$ \\
  \hline\hline
  \end{tabular}
}
\caption
{
  The performance comparison between CNN tracker and other $27$ trackers on the VOT2013
  benchmark. For each column, the best score is shown in bold and red while the second
  best score is shown in blue. 
}
\label{tab:tracking_resutls_vot_all}
\end{table}

Note that the scores listed in Table~\ref{tab:tracking_resutls_vot_all} and the plots in
Fig.~\ref{fig:vot_rank_plot} are rank-based, which is different from the measuring
criterion used in the CVPR2013 benchmark. It is well-known that the evaluation method for
visual tracker is not unique and could be sophisticated for a specific objective
\cite{nawaz2013protocol}. Usually different tracker measures offer different points of
view for accessing the tracking method. The best performance on the VOT2013 benchmark
justifies the superiority of DeepTrack, from another perspective.

In \cite{gao2014transfer}, the authors perform their TGPR tracker on the VOT2013
benchmark, without comparing with other trackers. We here compare our DeepTrack with the
TGPR algorithm, which is recently proposed and achieves state-of-the-art performance in
the CVPR2013 benchmark.
Following the settings in \cite{gao2014transfer}, we perform the proposed tracker in
Experiment-$1$ and Experiment-$2$. The performance comparison is shown in
Table~\ref{tab:tracking_resutls_vot}. 


\begin{table*}[htb]
\centering
\resizebox{1\textwidth}{!}
{
  \begin{tabular}{ l | c | c | c | c | c | c | c | c | c | c | c | c | c | c | c | c | c }
  \hline\hline
  & \emph{bicycle} & \emph{bolt} & \emph{car} & \emph{cup} & \emph{david} & \emph{diving}
  & \emph{face} & \emph{gym} & \emph{hand} & \emph{iceskater} & \emph{juice} & \emph{jump} 
  & \emph{singer} & \emph{sunshade} & \emph{torus} & \emph{woman} & overall \\
  \hline\hline
  Exp1-TPGR-Rob.  & $\bf{0}$ & $1.27$ & $\bf{0.40}$ & $\bf{0}$ & $0.27$ & $2.87$ & $\bf{0}$ & $2.87$ & $1.67$ &
           $\bf{0}$ & $\bf{0}$ & $\bf{0}$ & $0.60$ & $0.20$ & $0.13$ & $1.00$ & $0.71$ \\
  Exp1-DeepTrack-Rob.   & $0.47$ & $\bf{0.07}$ & $0.47$ & $0$ & $\bf{0.20}$ & $\bf{0.80}$ 
                  & $\bf{0}$ & $\bf{0.73}$ & $\bf{0.20}$ & $\bf{0}$ & $\bf{0}$ 
                  & $\bf{0}$ & $\bf{0}$ & $\bf{0}$ & $\bf{0.07}$ & $\bf{0.47}$ & $\bf{0.22}$ \\
  \hline
  Exp1-TPGR-Accu. & $0.60$ & $0.57$ & $0.45$ & $0.83$ & $0.58$ & $0.33$ & $0.85$ & $0.57$ & $0.56$ &
                    $0.60$ & $0.76$ & $0.59$ & $0.65$ & $0.73$ & $0.78$ & $0.74$ & $0.64$ \\
  Exp1-DeepTrack-Accu.  & $0.58$ & $0.61$ & $0.51$ & $0.86$ & $0.54$ & $0.35$ & $0.73$ & $0.49$
                  & $0.54$ & $0.61$ & $0.81$ & $0.66$ & $0.51$ & $0.72$ & $0.76$ & $0.60$
                  & $0.62$ \\
  \hline\hline
  
  Exp2-TPGR-Rob.  & $\bf{0}$ & $1.27$ & $\bf{0.20}$ & $\bf{0}$ & $0.27$ & $2.87$ & $0.07$ & $3.00$ & $2.07$ &
         $\bf{0}$ & $\bf 0$ & $\bf 0$ & $0.33$ & $\bf{0.07}$ & $0.60$ & $1.00$ & $0.73$ \\
  Exp2-DeepTrack-Rob.  & $0.27$ & $\bf{0}$ & $0.33$ & $\bf{0}$ & $\bf{0.20}$ 
                & $\bf{0.80}$ & $\bf{0}$ & $\bf{0.27}$ & $\bf{0.60}$ 
                & $\bf{0}$ & $\bf{0}$ & $\bf{0}$ & $\bf{0}$ & $0.07$ & $\bf{0.27}$ 
                & $\bf{0.67}$ & $\bf{0.22}$ \\
  \hline
  Exp2-TPGR-Accu. & $0.57$ & $0.57$ & $0.41$ & $0.75$ & $0.58$ & $0.32$ & $0.77$ & $0.53$ & $0.53$ &
                    $0.57$ & $0.73$ & $0.57$ & $0.45$ & $0.64$ & $0.65$ & $0.67$ & $0.58$ \\
  Exp2-DeepTrack-Accu.  & $0.54$ & $0.62$ & $0.49$ & $0.77$ & $0.50$ & $0.36$ & $0.70$ & $0.47$
                  & $0.53$ & $0.59$ & $0.75$ & $0.62$ & $0.60$ & $0.69$ & $0.69$ & $0.56$
                  & $0.59$ \\
  \hline\hline
  \end{tabular}
}
\caption
{
  The performance comparison between DeepTrack tracker and the TPGR tracker on the VOT2013
  benchmark. The better robustness score is shown in bold. Note that for accuracy (Accu.),
  the comparison is not fair if the robustness score is different and thus no bold
  accuracy score is shown.
}
\label{tab:tracking_resutls_vot}
\end{table*}

We can see that the proposed DeepTrack outperforms the TPGR tracker in the robustness
evaluation, with a clear performance gap. For Experiment-1, one needs to reinitialize
the TPGR tracker for $0.71$ times per sequence while that number for our method is only
$0.22$. Similarly, with the bounding box perturbation (Experiment-2), TPGR needs $0.73$
times re-initialization while DeepTrack still requires $0.22$ times. Note that in
Table~\ref{tab:tracking_resutls_vot} the accuracies from different trackers are not
directly comparable, as they are calculated based on different re-initialization
conditions. However, by observing the overall scores, we can still draw the conclusion
that the DeepTrack is more robust than TPGR as it achieves similar accuracies to TPGR ($0.62$ {\it
v.s.} $0.64$ for Experiment-1 and $0.59$ {\it v.s.} $0.58$ for Experiment-2) while only
requires around one third of re-initializations. 

\subsection{Verification for the structural loss and the robust temporal sampling}

\begin{figure*}[ht!]
  \centering
  \subfigure{\includegraphics[width=0.48\textwidth]{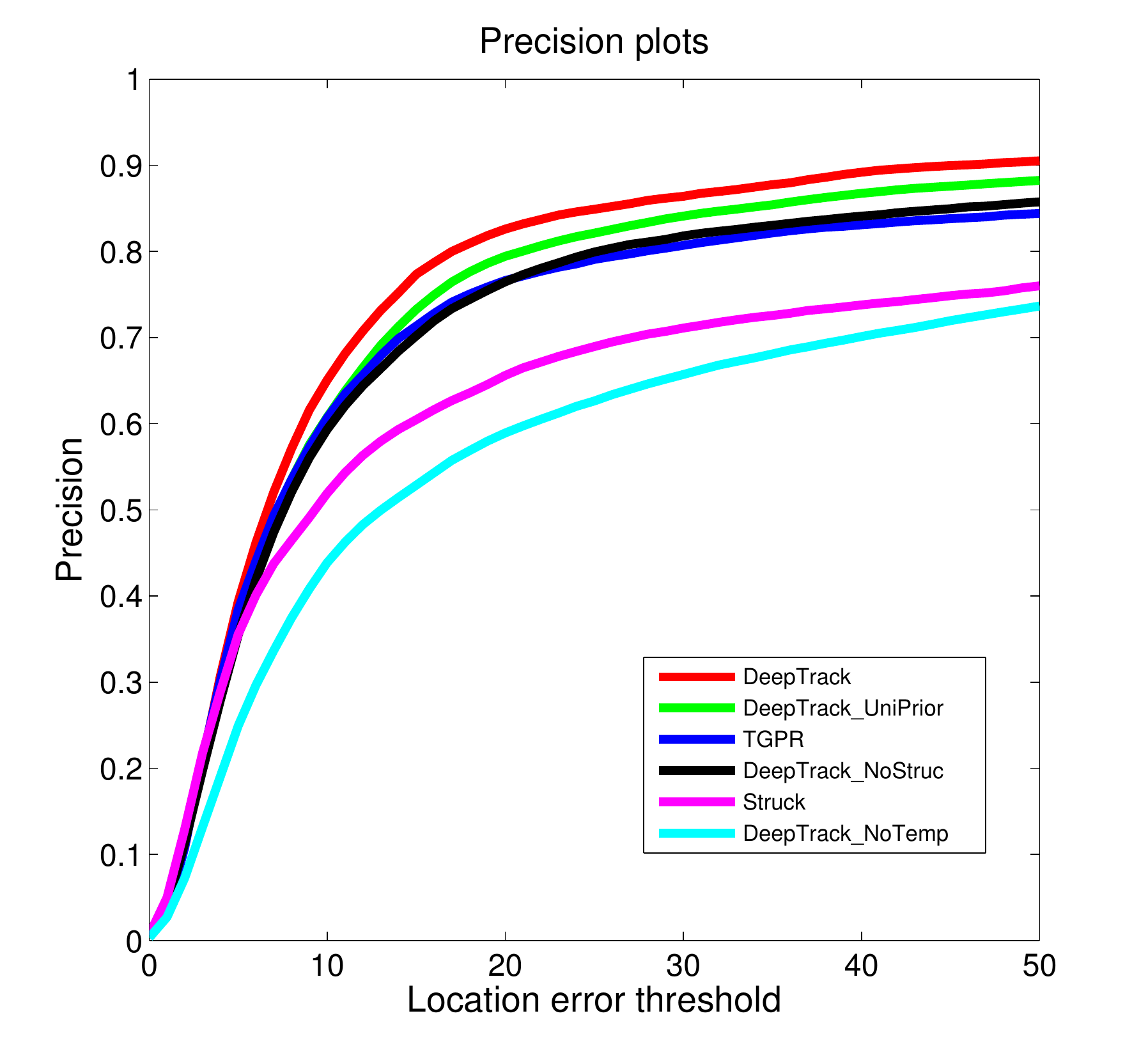}}
  \subfigure{\includegraphics[width=0.48\textwidth]{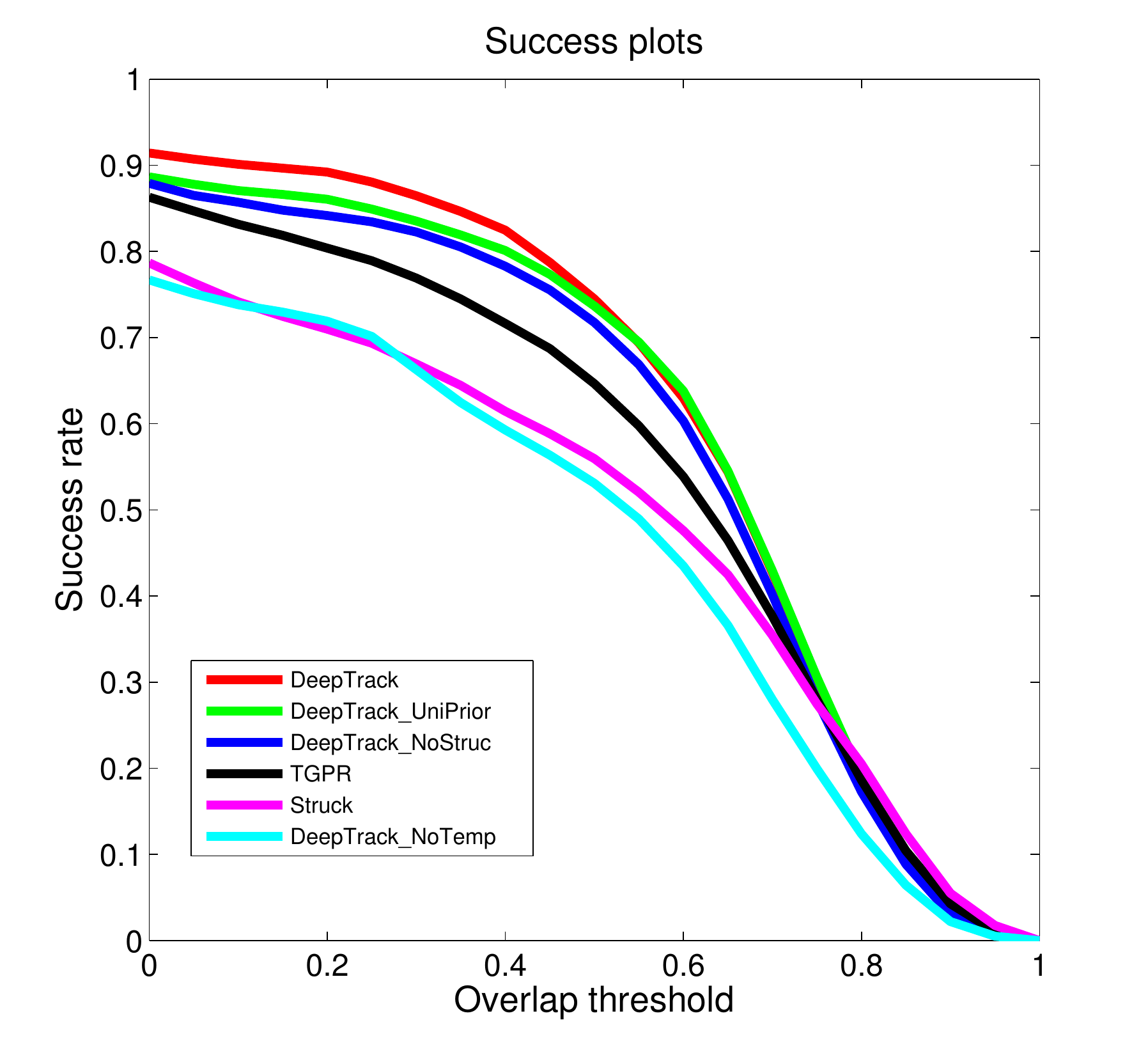}} \\
  \caption
  {
    The Precision Plot (left) and the Success Plot (right) of the results obtained by
    using different versions of DeepTrack. Note that the color of one curve is determined
    by the rank of the corresponding trackers, not their names. 
  }
\label{fig:pre_suc_class}%
\end{figure*}

Here we verify the three proposed modifications to the CNN model. We rerun the experiment
on the CVPR2013 benchmark using the DeepTrack with each modification inactivated. In
specific, the temporal sampling mechanism, the label uncertainty and the structural loss
is disabled and the yielded tracking results are shown in Fig.~\ref{fig:pre_suc_class},
compared with the full-version of the proposed method. Beside, the results of two
state-of-the-art method, {\emph i.e.}, Struck and TPGR are also shown as references.

From the figure we can see that, the structural loss, the temporal sampling mechanism and
the label uncertainty all contribute the success of our CNN tracker. In particular, the
temporal sampling plays a more important role. The structural loss can 
increase the TP accuracy by $10\%$ and one can lifts the TP accuracy by
$4\%$ when the label noise is taken into consideration. 
Generally speaking, the curve consistently goes down when one component are removed from
the original DeepTrack model. That indicates the validity of the propose modifications.

\subsection{Tracking speed analysis}
We report the average speed (in fps) of the proposed DeepTrack method in
Table~\ref{tab:speed}, compared with the DeepTrack without the truncated loss. Note that
there are two kinds of average speed scores: the average fps over all the sequences and
the average fps over all the frames. The latter one reduces the influence of short
sequences where the initialization process usually dominates the computational burden.


\begin{table}[htb]
\centering
{
  \begin{tabular}{| c | c | c |}

  \hline
  & Sequence Average & Frame Average  \\
  \cline{1-3}
  With TruncLoss & 1.96fps & 2.52fps \\
  \cline{1-3}
    No TruncLoss & 1.49fps & 1.86fps \\
  \hline
  \end{tabular}
}
\caption
{
  The tracking speed of DeepTrack with or without the truncated loss.  Note that there are
  two kinds of kinds of average speed scores: the average fps over all the sequences
  (Sequence Average) and the average fps over all the frames (Frame Average).
}
\label{tab:speed}
\end{table}

According to the table, the truncated loss boosts the tracking efficiency by around $37\%$.
Furthermore, our method tracks the object at an average speed around $2.5$fps.
Considering that the speed of TPGR is around $3$fps \cite{gao2014transfer} and for the
Sparse Representation based methods the speeds are usually lower than $2.5$fps
\cite{xing2013robust}. We thus can draw the conclusion that the DeepTrack can achieve
comparable speed to the state-of-the-art methods.

\section{Conclusion}

We introduced a CNN based online object tracker. We employed a novel CNN architecture and
a structural loss function that handles multiple input cues. We also proposed to modify
the ordinary Stochastic Gradient Descent for visual tracking by iteratively update the
parameters and add a robust temporal sampling mechanism in the mini-batch generation. This
tracking-tailored SGD algorithm increase the speed and the robustness of the training
process significantly. Our experiments demonstrated that the CNN-based DeepTrack
outperforms state-of-the-art methods on two recently proposed benchmarks which contain
over $60$ video sequences and achieves the comparable tracking speed.

\bibliographystyle{IEEEbib}
\bibliography{egbib_YL}
\end{document}